\documentclass[runningheads]{llncs}

\usepackage{eccv}
\usepackage{eccvabbrv}
\usepackage{graphicx}
\usepackage{booktabs}

\usepackage{amsfonts}
\usepackage{nicefrac}
\usepackage{microtype}
\usepackage{adjustbox}
\usepackage{nicematrix}
\usepackage{marvosym}
\usepackage{multirow}
\usepackage{tabularx}
\newcolumntype{C}{>{\centering\arraybackslash}X}

\usepackage[accsupp]{axessibility}

\usepackage{hyperref}

\usepackage{orcidlink}

\usepackage[dvipsnames, table]{xcolor}
\renewcommand{\thefootnote}{}
\newcommand{\ourmethod}{RT-DETRv4}

\definecolor{cvprblue}{rgb}{0.21,0.49,0.74}
\definecolor{aliceblue}{rgb}{0.94,0.97,1.0}

\makeatletter
\DeclareRobustCommand{\equalmark}{\@fnsymbol{1}}
\makeatother

\begin{document}

\title{RT-DETRv4: Painlessly Furthering Real-Time Object Detection with Vision Foundation Models}

\titlerunning{RT-DETRv4}

\author{Zijun Liao\inst{1}\thanks{Equal contribution. \dag Project leader. \textrm{\Letter} Corresponding authors.}\orcidlink{0009-0004-7232-5274} \and
Yian Zhao\inst{1\equalmark\dag}\orcidlink{0000-0001-7943-3164} \and
Xin Shan\inst{1}\orcidlink{0009-0007-4200-6661} \and
Yu Yan\inst{1}\orcidlink{0009-0004-7466-7083} \and
Chang Liu\inst{5\textrm{\Letter}}\orcidlink{0000-0001-6747-0646} \and
Lei Lu\inst{1}\orcidlink{0000-0001-5275-2181} \and
Xiangyang Ji\inst{5\textrm{\Letter}}\orcidlink{0000-0002-7333-9975} \and
Jie Chen\inst{4,2,1,3\textrm{\Letter}}\orcidlink{0000-0002-9765-4523}}

\authorrunning{Z.~Liao et al.}

\institute{School of Electronic and Computer Engineering, Peking University, Shenzhen, China
\and
Pengcheng Laboratory, Shenzhen, China
\and
AI for Science (AI4S)-Preferred Program, Peking University Shenzhen Graduate School, China
\and
School of Intelligence Science and Engineering, Harbin Institute of Technology, Shenzhen, China
\and
Department of Automation and BNRist, Tsinghua University, Beijing, China \\
\email{jiechen2019@pku.edu.cn}
% \email{\{zjliao25,zhaoyian\}@stu.pku.edu.cn}
}

\maketitle
\setcounter{figure}{-1}

\begin{center}
\vspace{-0.55cm}
    \captionsetup{type=figure}
    \hfill
    \begin{subfigure}{0.494\textwidth}
        \includegraphics[width=0.98\textwidth]{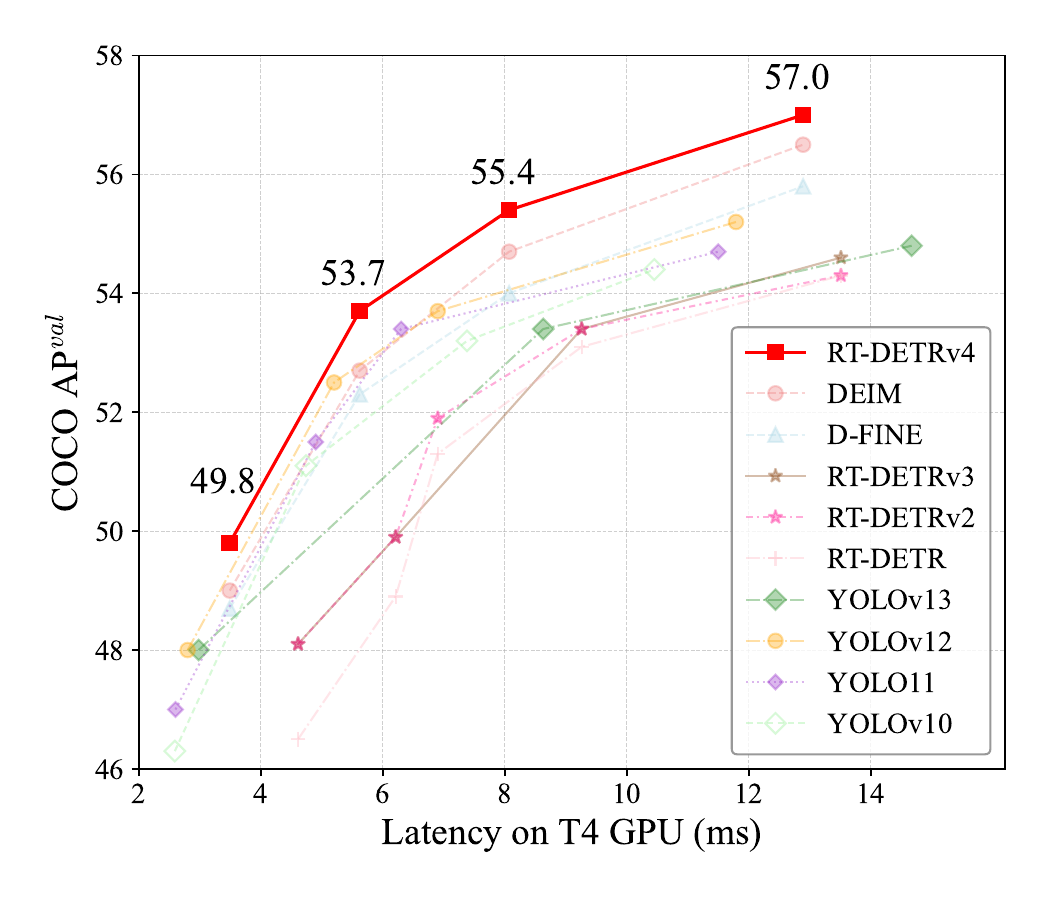}
        \vspace{-0.15cm}
        \caption{\textbf{COCO AP$^{val}$ vs. T4 GPU Latency.}}
        \label{fig:coco_ap_latency}
        
    \end{subfigure}
    \hfill
    \begin{subfigure}{0.494\textwidth}
        \includegraphics[width=0.98\textwidth]{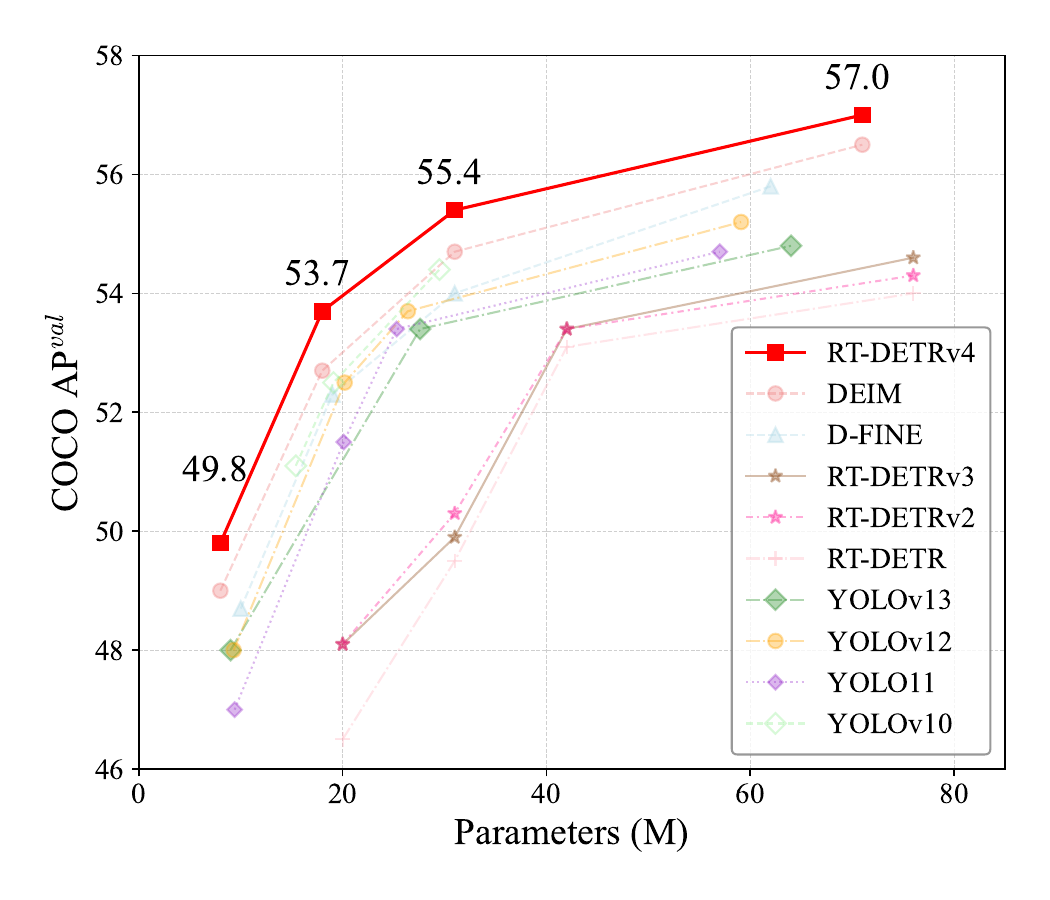}
        \vspace{-0.15cm}
        \caption{\textbf{COCO AP$^{val}$ vs. Params (M).}}
        
    \end{subfigure}
    \hfill 
    \vspace{-0.6cm}
    \captionof{figure}{
    \textbf{Compared with advanced real-time object detectors on COCO \cite{lin2014microsoft}}. 
    % Our RT-DETRv4 models achieve state-of-the-art performance.
    }
    \vspace{-0.2cm}
    \label{fig:teaser}
\end{center}

\begin{abstract}
Real-time object detection has achieved substantial progress through meticulously designed architectures and optimization strategies. 
However, the pursuit of high-speed inference via lightweight network designs often leads to degraded feature representation, which hinders further performance improvements and practical on-device deployment. 
In this paper, we propose a cost-effective and highly adaptable distillation framework that harnesses the rapidly evolving capabilities of Vision Foundation Models (VFMs) to enhance lightweight object detectors. 
Given the significant architectural and learning objective disparities between VFMs and resource-constrained detectors, achieving stable and task-aligned semantic transfer is challenging. To address this, on one hand, we introduce a \textbf{Deep Semantic Injector (DSI)} module that facilitates the integration of high-level representations from VFMs into the deep layers of the detector. On the other hand, we devise a \textbf{Gradient-guided Adaptive Modulation (GAM)} strategy, which dynamically adjusts the intensity of semantic transfer based on gradient norm ratios. Without increasing deployment and inference overhead, our approach painlessly delivers striking and consistent performance gains across diverse DETR-based models, underscoring its practical utility for real-time detection. 
Our new model family, RT-DETRv4, achieves state-of-the-art results on COCO, attaining AP scores of $49.8/53.7/55.4/57.0$ at corresponding speeds of $273/169/124/78$ FPS. Code is publicly available at \url{https://github.com/RT-DETRs/RT-DETRv4}.
\end{abstract}
\section{Introduction}
\label{sec:intro}
Real-time object detection stands as a fundamental task in computer vision, which underpins numerous interactive and safety-critical applications that demand instant perception and decision making, such as autonomous driving~\cite{chen2024end}, embodied intelligence~\cite{liu2025embodied}, and human–computer interaction~\cite{kirillov2023segment, zhao2024graco, ravi2025sam}. 
Over the past decade, remarkable progress has been driven by efficient network architectures and end-to-end learning frameworks. 
In particular, two representative series, YOLO~\cite{redmon2016you} and DETR~\cite{carion2020end}, have profoundly influenced the evolution of object detection paradigms. The YOLO family emphasizes rapid one-stage detection achieving high inference speed and practical deployment efficiency, and the DETR series has reshaped the detection paradigm through its unified modeling of object queries and set-based prediction.
Among its variants, RT-DETR~\cite{zhao2024detrs} marked a milestone as the first real-time DETR, introducing the DETR family to the real-time community by outperforming YOLO models in both speed and accuracy.

Despite the notable progress, a long-standing challenge remains: the inherent trade-off between designing lightweight models to achieve high inference speed and employing complex architectures to improve feature representation.
To meet real-time constraints, detectors typically adopt lightweight backbones and carefully designed computational modules, which inevitably reduce their ability to capture high-level semantics and lead to a semantic bottleneck.
This limitation not only hinders further performance improvement, but also increases the difficulty of practical on-device deployment.

In this paper, inspired by the rapid advances in Vision Foundation Models (VFMs)~\cite{simeoni2025dinov3,he2022masked}, we propose a cost-effective and highly adaptable distillation framework that leverages the powerful representational capacity of VFMs to enhance lightweight object detectors.
By transferring the rich semantics of VFMs to real-time detectors during training while keeping the detector architecture unchanged during inference, our method enables significant enhancement without introducing any additional inference or deployment cost.
This advantage is particularly important for practical real-time detection applications.

However, achieving stable and task-aligned semantic transfer is challenging due to the large architectural and learning objective disparities between VFMs and resource-constrained detectors.
To address this issue, we first introduce a Deep Semantic Injector (DSI) module that enables the integration of high-level representations from VFMs into the deep layers of the detector.
To ensure stable and efficient optimization, we further design a Gradient-guided Adaptive Modulation (GAM) strategy that dynamically adjusts the strength of semantic injection based on gradient norm ratios, thereby harmonizing the learning of semantic transfer and detection objectives.

Extensive experiments show that the proposed method achieves significant and consistent performance improvements over advanced DETR-based detectors without increasing inference or deployment overhead, underscoring its effectiveness.
In summary, our main contributions are as follows:
\begin{itemize}
\item We propose a cost-effective and highly adaptable distillation framework that leverages the evolving capabilities of VFMs to painlessly enhance real-time detectors, providing a scalable pathway for transferring foundation-level semantics to lightweight architectures.
\item We propose the Deep Semantic Injector (DSI) and Gradient-guided Adaptive Modulation (GAM), which enable stable and task-aligned semantic transfer between VFMs and detectors with significantly different architectures and learning objectives.
\item We establish a new family of models, \ourmethod-S/M/L/X, achieving $49.8 / 53.7 / 55.4 / 57.0$ AP scores on COCO~\cite{lin2014microsoft} at $273 / 169 / 124 / 78$ FPS, setting a new SOTA on COCO dataset.
\end{itemize}
\section{Related work}
\label{sec:related_work}

\subsection{Real-time object detection}
The evolution of real-time object detection has long been driven by the You Only Look Once (YOLO) family~\cite{redmon2016you}, which popularized the single-stage paradigm through an efficient and unified detection pipeline. Over the past few years, this lineage has undergone rapid iteration, introducing continuous refinements in backbone design, label assignment, and optimization strategy~\cite{redmon2017yolo9000, redmon2018yolov3, bochkovskiy2020yolov4, yolov5v7.0, li2022yolov6, wang2023yolov7, yolov8, wang2024yolov9}.
Recent generations have expanded the design space even further: YOLOv10~\cite{wang2024yolov10} eliminated NMS that the YOLO series has long relied on, YOLO11~\cite{yolo11} improved the architectural hierarchy and neck connectivity, YOLOv12~\cite{tian2025yolov12} incorporated attention mechanisms for better contextual reasoning, and YOLOv13~\cite{lei2025yolov13} explored hypergraph representations to capture higher-order feature dependencies.
These advances have pushed the performance–efficiency frontier of convolutional and hybrid architectures, gradually narrowing the gap between real-time and high-accuracy detectors.

In parallel, another line of research has evolved around the DEtection TRansformer (DETR)~\cite{carion2020end}, which redefined object detection as a set prediction problem and eliminated hand-crafted components such as anchor design and NMS. This transformer-based paradigm inspired numerous variants, including Deformable DETR~\cite{zhu2021deformable}, Conditional DETR~\cite{meng2021conditional}, and DAB-DETR~\cite{liu2022dabdetr}, which focus on improving convergence and localization accuracy. Later works such as DN-DETR~\cite{li2022dn}, DINO~\cite{zhang2023dino}, and Group-DETR~\cite{chen2023group} introduced denoising objectives and group-wise supervision to further enhance training stability and representational quality.

Building on this foundation, RT-DETR~\cite{zhao2024detrs} established the first real-time end-to-end transformer detector that achieved parity with, and in some cases surpassed, contemporary YOLO models. Subsequent works have continued to improve its training efficiency and representation learning without incurring inference overhead. For instance, RT-DETRv2~\cite{lv2024rt} and RT-DETRv3~\cite{wang2025rt} incorporated auxiliary supervision for enhanced gradient flow, D-FINE~\cite{peng2025d} employed self-distillation to refine semantic representation, and DEIM~\cite{huang2025deim} introduced dense matching for more precise feature alignment.
Collectively, these developments illustrate a clear trend: as architectural efficiency saturates, training supervision and semantic representation become the primary levers for further progress. Our work builds on this insight by strengthening the core representation via deep semantic transfer, achieving higher accuracy at no additional deployment or inference cost.

\subsection{Knowledge distillation for object detection}
Knowledge distillation has been widely explored to improve object detectors under efficiency constraints. FGD~\cite{yang2022focaland} transfers focal local knowledge and global relational information by modeling the uneven importance of regions and channels. DETRDistill~\cite{chang2023detrdistill} adapts distillation to DETRs via Hungarian-matching logits, target-aware feature, and query-prior assignments. CrossKD~\cite{wang2024crosskd} forwards the student's head features through the teacher head for more task-aligned prediction distillation. These studies show that customized supervision can enhance detector representations without additional inference cost. However, most methods rely on teachers trained for the target detection task, whereas Vision Foundation Models offer more general semantic priors, motivating our use of a frozen VFM as an external teacher.

\subsection{Vision Foundation Models}
Vision Foundation Models (VFMs) have become a dominant paradigm for learning general-purpose vision representations from large-scale image corpora with minimal or no human supervision. Early progress stemmed from self- and weakly-supervised learning methods, which enabled models to capture high-level semantics from unlabeled or loosely labeled data. Representative approaches include contrastive learning frameworks such as SimCLR~\cite{chen2020simple} and MoCo~\cite{he2020momentum}, which learn discriminative features by enforcing consistency across augmented views of the same image while contrasting them with others. CLIP~\cite{radford2021learning} further extended this idea to large-scale image–text contrastive training, aligning visual and linguistic embeddings and demonstrating strong zero-shot transferability across diverse tasks.

Inspired by masked language modeling in NLP, Masked Image Modeling (MIM) approaches were introduced to reconstruct masked image regions, thereby learning context-aware and holistic representations. Notable methods include MAE~\cite{he2022masked} and BEiT~\cite{bao2022beit}. In parallel, I-JEPA~\cite{assran2023self} learns semantic image representations with a joint embedding predictive objective, predicting target block representations from a context block. The DINO family~\cite{caron2021emerging, oquab2023dinov2, simeoni2025dinov3} integrates contrastive, reconstruction-based, and self-distillation objectives to produce highly semantic and transferable features. In particular, DINOv3~\cite{simeoni2025dinov3} demonstrates the scalability and efficacy of large-scale self-supervised learning, achieving rich and robust representations without human annotations.
\begin{figure*}[t]
    \centering
    \includegraphics[width=1\linewidth]{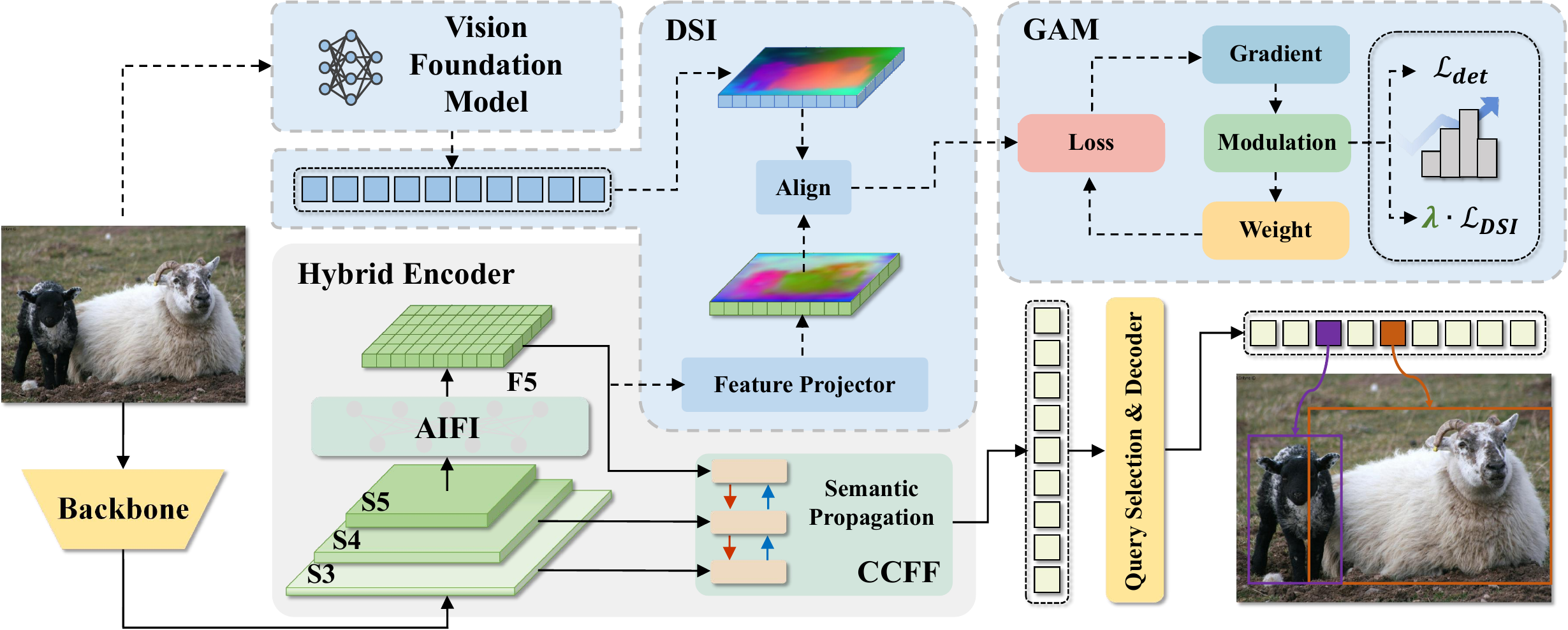}
    \caption{\textbf{Overview of RT-DETRv4}. We leverage a Vision Foundation Model (VFM) to extract high-quality semantic representations, which are aligned with the deepest feature map ($F_5$) from the AIFI module via a Feature Projector in the Deep Semantic Injector (DSI). To ensure faster and more stable convergence, a Gradient-guided Adaptive Modulation (GAM) dynamically adjusts the DSI loss during training. The proposed framework operates only during the training phase (dashed arrows and blue blocks) of the real-time detector and keeps the original architecture unchanged during inference and deployment, introducing no additional overhead while improving accuracy.}
    \label{fig:overview}
\end{figure*}

\section{Method}
\label{sec:method}

\subsection{Motivation}
\label{sec:motivation}
\begin{figure*}[t]
    \centering
    \includegraphics[width=1\linewidth]{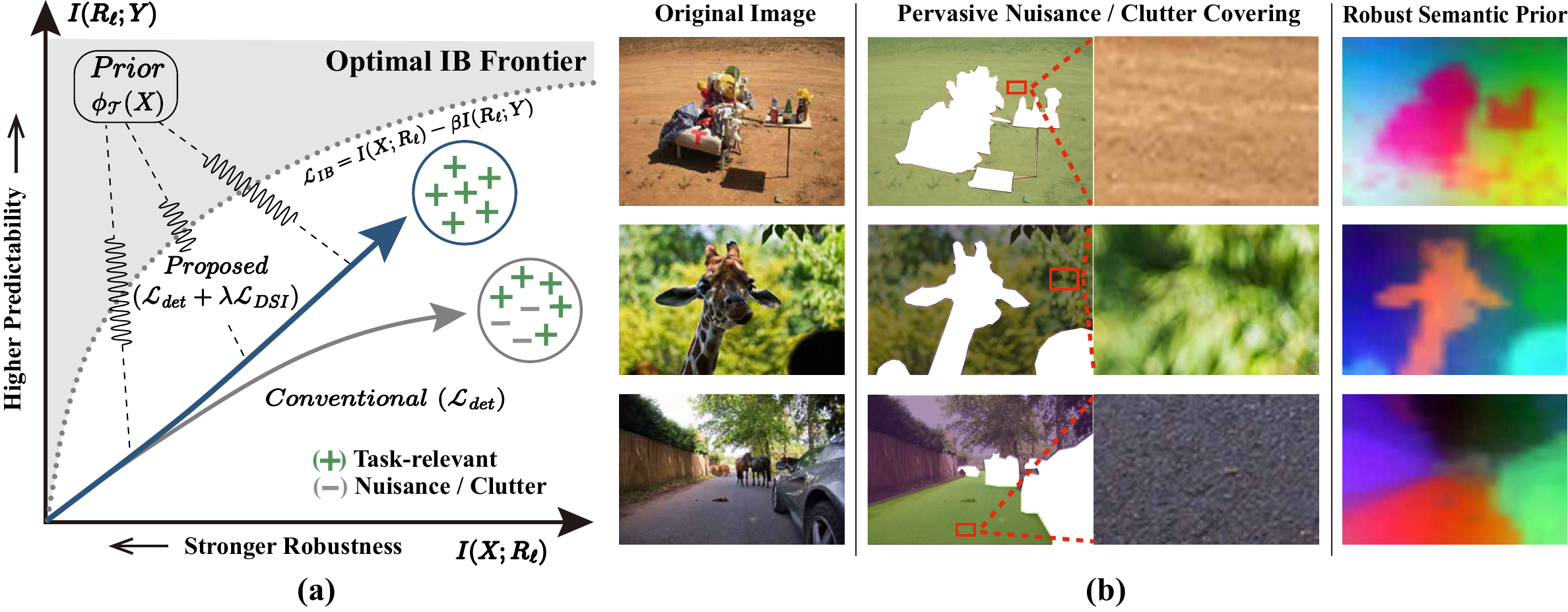}
    \caption{\textbf{Semantic injection for object detection.} 
    (a) Illustration from the information bottleneck perspective. 
    (b) Visualization of nuisance details in images versus robust semantic representations from a VFM. 
    }
    \label{fig:intuition}
\end{figure*}

\paragraph{Information Bottleneck for detection representations.} 
According to the Information Bottleneck (IB) principle~\cite{tishby2000information}, optimal representations should preserve task-relevant semantics while discarding nuisance from the input.
Let $X$ denote the input image, $Y$ denote the detection target, and $I(\cdot;\cdot)$ denote mutual information; optimal representations $R_\ell$ on the $\ell$-th scale should minimize
\begin{equation}
\mathcal{L}_{IB}=I(X;R_\ell)-\beta I(R_\ell;Y),
\label{eq:information_bottleneck}
\end{equation}
where $\beta$ controls the trade-off between the compression of the input space and the predictive accuracy for the target domain.
However, almost all detectors are optimized solely through task-driven supervision, i.e., by maximizing predictive mutual information $I(R_\ell; Y)$ via back-propagation from object detection loss:
\begin{equation}
\mathcal{L}_{det} \sim - I(R_\ell; Y), 
\label{eq:loss_det_information_bottleneck}
\end{equation}
as indicated by the gray curve in~\cref{fig:intuition}(a). This training paradigm implicitly encourages representations that are discriminative for $Y$, but lacking an explicit constraint on $I(X; R_\ell)$. There is no direct mechanism to suppress non-robust components inherited from $X$, such as background clutter or texture-level patterns that are weakly correlated with object semantics, as illustrated in~\cref{fig:intuition}(b).

\paragraph{Semantic supervision from VFMs.}
Under the IB principle in~\cref{eq:information_bottleneck}, a desirable representation should preserve task-relevant information while discarding nuisance factors. To this end, we introduce semantic supervision from a pretrained VFM, which empirically serves as a strong and robust semantic prior that can be viewed as semantically more compressed, emphasizing high-level object semantics while being less sensitive to low-level nuisance factors:
\begin{equation}
T = \phi_{\mathcal{T}}(X),
\quad 
I(X; T) \lesssim I\!\left(X; R_\ell\right),
\label{eq:teacher_representation}
\end{equation}
Such representations are more robust to appearance changes and background noise.
We therefore introduce semantic supervision during detector training:
\begin{equation}
\mathcal{L}=\mathcal{L}_{det}+\lambda \mathcal{L}(R_\ell, T).
\label{eq:loss_semantic}
\end{equation}
as indicated by the blue curve in~\cref{fig:intuition}(a). The detection loss encourages task-specific discrimination, while the semantic objective aligns detector features with robust high-level semantics. The coupled objective consequently improves detection performance. Meanwhile, robust features benefit model generalization and help reduce overfitting.

\paragraph{Principle of Effective Semantic Injection.}
Most modern detectors follow the \textit{Backbone--Neck--Head} architecture, where the feature extraction process can be modeled as
\begin{equation}
X \xrightarrow{\text{Backbone}} \{S_\ell\} \xrightarrow{\text{Neck}} \{F_\ell\} \xrightarrow{\text{Head}} \hat{Y},\quad R_\ell \in \{S_\ell, F_\ell\}.
\end{equation}
According to the Data Processing Inequality,
\begin{equation}
I(X;S_1) \ge I(X;S_2) \ge \cdots \ge I(X;F_1) \ge I(X;F_2) \ge \cdots .
\end{equation}
Thus deeper layers naturally perform progressive information compression and should encode increasingly abstract semantics. Consequently, injecting semantic supervision at relatively deeper stages is theoretically preferable.

\begin{figure*}[t]
    \centering
    \includegraphics[width=1\linewidth]{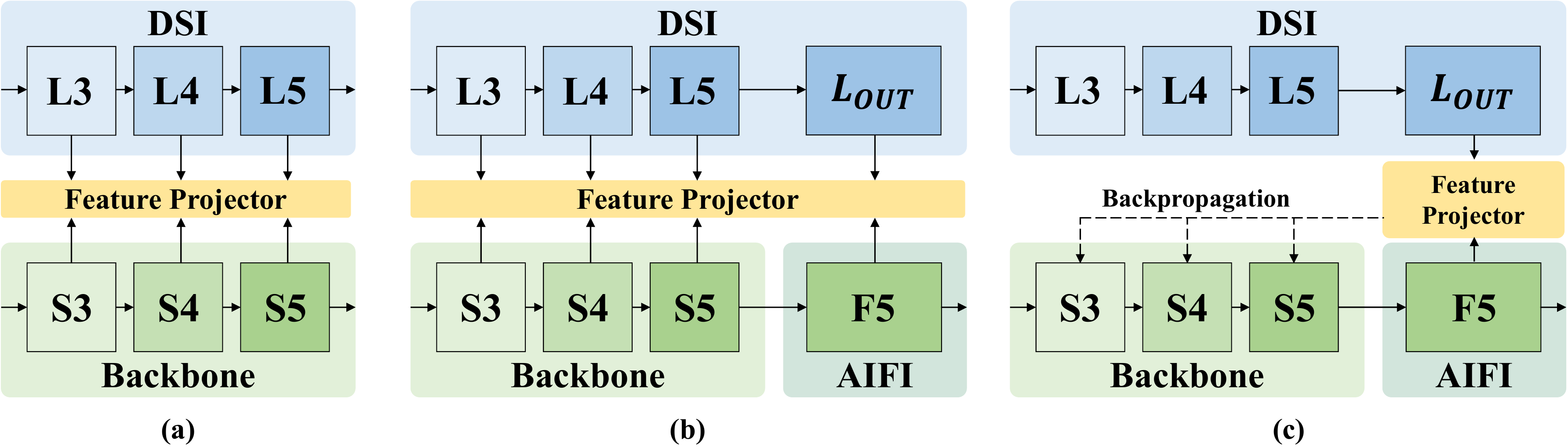}
    \caption{\textbf{Illustration of different Deep Semantic Injector (DSI) strategies.} 
    (a) Direct alignment of multi-scale backbone features (\(S_3, S_4, S_5\)). 
    (b) Hybrid alignment of both backbone and the AIFI features (\(S_3, S_4, S_5, F_5\)). 
    (c) Ours: Targeted alignment of only the AIFI feature (\(F_5\)), which possesses the highest-level semantics. This design allows gradients to backpropagate, enhancing both the AIFI module and the backbone.
    }
    \label{fig:module_design}
\end{figure*}

\subsection{Overview}
We focus on applying our framework to DETR-based real-time object detectors, \textit{i.e.}, RT-DETR models~\cite{zhao2024detrs}.
The overall framework of our method is shown in~\cref{fig:overview}, where the proposed modules are highlighted in blue. 
Given multi-scale backbone features \((S_3,S_4,S_5)\), RT-DETR uses self-attention only on the deepest map \(S_5\) via the Attention-based Intra-scale Feature Interaction (AIFI), producing an enhanced deep feature \(F_5\).
Then CNN-based Cross-scale Feature Fusion (CCFF) propagates information from \(F_5\) to obtain multi-scale outputs for the decoder.

The design of the hybrid encoder makes the quality of the feature map $F_5$ particularly critical. 
As the only feature subjected to self-attention, $F_5$ serves as the principal source of high-level, global semantic information for the entire model. Its quality directly affects the subsequent cross-scale fusion in the CCFF module, the initial query selection, and ultimately the performance of the decoder. 
However, the AIFI module that produces $F_5$ is trained with only indirect supervision, as the gradients from the final detection losses must backpropagate through the decoder and CCFF before reaching $F_5$. Such indirect supervision may be insufficient to fully optimize $F_5$. This leads to the major \textit{Semantic Bottleneck} in RT-DETR. Therefore, we prioritize performing semantic injection on $F_5$, while also considering alternative designs involving other feature levels and mixed injection strategies, as illustrated in~\cref{fig:module_design}.

Based on the discussion in~\cref{sec:motivation}, we propose the \textbf{Deep Semantic Injector (DSI)}, a lightweight training-only module that explicitly aligns the deep feature with semantically rich representations from a VFM.
This targeted supervision enhances the semantic expressiveness of high-level features and allows the gradient to flow back through AIFI and the backbone, improving both modules synergistically.
With DSI incorporated, the total training objective is defined as:
\begin{equation}
\mathcal{L}_{\text{total}} = \mathcal{L}_{\text{det}} + \lambda \mathcal{L}_{\text{DSI}},
\label{eq:loss_all}
\end{equation}
where $\mathcal{L}_{\text{det}}$ denotes the standard detection loss (e.g., classification and bounding box regression), and $\mathcal{L}_{\text{DSI}}$ represents the proposed semantic alignment loss.
However, achieving stable and task-aligned semantic transfer is challenging because of the large architectural and learning objective disparities between VFMs and resource-constrained detectors.
An inappropriate choice of $\lambda$ may either provide insufficient semantic supervision in the early stages or excessively dominate the detection objective in later stages, ultimately impeding convergence and degrading performance.
To adapt to the evolving optimization dynamics during training, we propose \textbf{Gradient-guided Adaptive Modulation (GAM)}, a mechanism that dynamically tunes $\lambda$ based on gradient statistics, ensuring balanced optimization between detection and semantic supervision.

\subsection{Deep Semantic Injector}
\label{subsec:dsi}

To address the \textit{Semantic Bottleneck}, we introduce the Deep Semantic Injector (DSI), a training-only module designed to provide explicit and powerful semantic supervision of $R_\ell$ by aligning it with representations from a high-capacity semantic teacher, denoted as $\mathcal{T}$. Given an input image $X$, $\mathcal{T}$ produces a high-quality feature representation $\phi_{\mathcal{T}}(X)$.

\paragraph{Feature projector.}
To align the detector's feature map $R_\ell \in \mathbb{R}^{H_\ell \times W_\ell \times C_k}$ with the teacher's representation, differences in both spatial resolution and channel dimensionality must be reconciled.
The teacher, typically a ViT~\cite{dosovitskiy2021an}, outputs a sequence of patch tokens $T_p \in \mathbb{R}^{N_p \times C_{\mathcal{T}}}$. To enable spatial comparison, $T_p$ is reshaped into a 2D grid representation $T_{sp} \in \mathbb{R}^{H_{\mathcal{T}} \times W_{\mathcal{T}} \times C_{\mathcal{T}}}$, where $N_p = H_{\mathcal{T}} \times W_{\mathcal{T}}$.
We then introduce a lightweight feature projector $\mathcal{P}$ to achieve twofold alignment. First, $T_{sp}$ is interpolated to match the spatial resolution of $R_\ell$. Meanwhile, $\mathcal{P}$ adjusts the channel dimension of $R_\ell$ to align with the teacher's semantic space. 
The complete projection process is summarized as follows:
\begin{equation}
T_{sp} = \text{Reshape}(T_p),
\quad 
T_{sp}' = \text{Interpolate}(T_{sp}), 
\quad 
R_\ell' = \mathcal{P}(R_\ell).
\label{eq:dsi}
\end{equation}

\paragraph{Alignment loss.} 
To encourage the detector to capture the rich semantics of the teacher's representation, we adopt a cosine similarity loss.
We maximize the patch-wise cosine similarity between the detector's projected features $R_\ell'$ and the teacher's projected features $T_{sp}'$, formulated as minimizing the negative cosine similarity averaged over all spatial locations $(i, j)$:
\begin{equation}
  \mathcal{L}_{DSI}(R_\ell', T_{sp}') = - \frac{1}{H_\ell W_\ell} \sum_{i,j} \frac{R_\ell'(i,j) \cdot T_{sp}'(i,j)}{\|R_\ell'(i,j)\| \|T_{sp}'(i,j)\|}.
  \label{eq:cos_sim}
\end{equation}

\paragraph{Semantic injection.}
As illustrated in~\cref{fig:module_design}, we design three progressively enhanced configurations for semantic injection. 
In configurations (a) and (b), the DSI module aligns features at different hierarchical depths via the Feature Projector, injecting semantic knowledge from the frozen VFM into the detector’s feature hierarchy. 
In configuration (c), given the pivotal role of the AIFI module in the hybrid encoder, alignment is applied to its output $F_5$, which contains the richest semantics. 
Without gradient detachment, the DSI loss backpropagates to update the lightweight backbone, while the forward pass leverages the enriched $F_5$ to guide cross-scale fusion in the CCFF module. 
This dual-directional supervision unifies semantic enhancement across the detector. 
The comparative results of different configurations~(\textit{cf.}~\cref{subsec:ablation}) demonstrate that configuration (c) achieves the best performance.

\subsection{Gradient-guided Adaptive Modulation}
\label{subsec:gam}
To ensure stable and adaptive semantic supervision, we propose a dynamic Gradient-guided Adaptive Modulation (GAM) mechanism that regulates the DSI loss weight ($\lambda$) based on the relative gradient contribution of the target module of semantic injection according to its \textit{gradient norm ratio}. This gradient-based regulation adaptively maintains balanced optimization.

Specifically, for each training step $t$ within epoch $e$, we compute the $L_1$ norm of gradients for each component $\mathcal{C}$:
\begin{equation}
  G_t^{(\mathcal{C})} = \| \nabla_{\theta_\mathcal{C}} \mathcal{L}_{total} \|_1.
% , \quad k \in \{\text{backbone}, \text{AIFI}, \text{CCFF}, \text{encoder\_others}, \text{decoder}\}.
  \label{eq:grad}
\end{equation}
The total gradient magnitude is given by:
\begin{equation}
  G_t^{(total)} = \sum_\mathcal{C} G_t^{(\mathcal{C})},
  \label{eq:grad_total}
\end{equation}
and the relative gradient contribution of the target module is defined as:
\begin{equation}
  r_t = \frac{G_t^{(\text{target})}}{G_t^{(total)}}.
  \label{eq:aifi_ratio}
\end{equation}
Then average the gradient ratios across all training steps within epoch $e$ to obtain:
\begin{equation}
  \bar{r}_e = \frac{1}{T_e} \sum_{t=1}^{T_e} r_t,
  \label{eq:avg_aifi_ratio}
\end{equation}
where $T_e$ denotes the number of steps in epoch $e$.

At the end of each epoch, GAM checks whether $\bar{r}_e$ lies within the target interval.
If $\bar{r}_e \in [\rho - \delta, \rho + \delta]$, the weight $\lambda_e$ of $\mathcal{L}_{\text{DSI}}$ remains unchanged.
Otherwise, $\lambda_{e}$ is adjusted such that the next epoch's gradient ratio of the target module of semantic injection is steered toward the \textit{further boundary} of the target range rather than its midpoint, since only a portion of gradients originates from $\mathcal{L}_{\text{DSI}}$, boundary-based adjustment yields more stable convergence near equilibrium.

\begin{equation}
    \lambda_{e+1} =
    \begin{cases}
    \lambda_e \cdot \dfrac{\rho - \delta}{\bar{r}_e}, & \text{if } \bar{r}_e > \rho + \delta, \\[8pt]
    \lambda_e \cdot \dfrac{\rho + \delta}{\bar{r}_e}, & \text{if } \bar{r}_e < \rho - \delta, \\[8pt]
    \lambda_e, & \text{otherwise.}
    \end{cases}
    \label{eq:update_method}
\end{equation}

Two hyperparameters govern the modulation process: the target ratio ($\rho$) defines the desired supervision intensity, whereas the tolerance interval ($\delta$) regulates the trade-off between responsiveness and stability.
This update rule drives the effective gradient contribution of semantic injection to converge within the desired operational range while preventing oscillations.
In practice, GAM provides stable convergence and consistently improves semantic alignment without additional tuning overhead.
\section{Experiments}

\subsection{Comparison with SOTA}
We compare our proposed {RT-DETRv4} with recent state-of-the-art real-time detectors, including the latest YOLO series (YOLOv10~\cite{wang2024yolov10}, YOLO11~\cite{yolo11}, YOLOv12~\cite{tian2025yolov12}, and YOLOv13~\cite{lei2025yolov13}) and DETR-based detectors (RT-DETR~\cite{zhao2024detrs}, RT-DETRv2~\cite{lv2024rt}, RT-DETRv3~\cite{wang2025rt}, D-FINE~\cite{peng2025d}, and DEIM~\cite{huang2025deim}). The results are illustrated in Fig.~\ref{fig:teaser}, and detailed statistics are provided in Tab.~\ref{tab:comparison}.
The results demonstrate that {RT-DETRv4} consistently achieves the best performance across all model scales (S, M, L, and X) without introducing any extra inference and deployment overhead.

Specifically, our \textbf{RT-DETRv4-L} achieves \textbf{55.4 AP} on COCO at \textbf{124 FPS}, outperforming YOLOv13-L (53.4 AP) and DEIM-L (54.7 AP) under comparable or even lower computational budgets.
The largest variant, \textbf{RT-DETRv4-X}, reaches \textbf{57.0 AP}, exceeding DEIM-X (56.5 AP) without introducing any inference overhead.
At smaller scales, \textbf{RT-DETRv4-S} and \textbf{RT-DETRv4-M} obtain \textbf{49.8} and \textbf{53.7 AP}, clearly surpassing their DEIM counterparts (49.0 and 52.7 AP). 

\begin{table}[!t]
    \caption{\textbf{Comparison with real-time object detectors on COCO~\cite{lin2014microsoft} \texttt{val2017}}. Results are sourced from the official publications. Unreported values evaluated from public weights using the standard protocol are marked with $^*$. R18, R34, R50, and R101 denote ResNet-18, ResNet-34, ResNet-50, and ResNet-101, respectively.} 
    \label{tab:comparison}
    \vspace{-5mm}
    \begin{center}
    \begin{adjustbox}{width=\textwidth}
    \begin{tabular}{l | cccc | ccc ccc}
        \toprule
        \textbf{Model} & \textbf{\#Epochs} & \textbf{\#Params.} & \textbf{GFLOPs} & \textbf{Latency (ms)} & \textbf{AP$^{val}$} & \textbf{AP$^{val}_{50}$} & \textbf{AP$^{val}_{75}$} & \textbf{AP$^{val}_S$} & \textbf{AP$^{val}_M$} & \textbf{AP$^{val}_L$} \\
        \midrule

        YOLOv10-S~\cite{wang2024yolov10} & 500 & 7 & 22 & 2.52 & 46.3 & 63.0 & 50.4 & 26.8 & 51.0 & 63.8 \\
        YOLO11-S~\cite{yolo11} & 500 & 9 & 22 & 2.60 & 47.0 & 63.4$^*$ & 50.5$^*$ & - & - & -\\
        YOLOv12-S~\cite{tian2025yolov12} & 600 & 9 & 21 & 2.78 & 48.0 & 65.0 & 51.8 & 29.8 & \underline{53.2} & 65.6 \\
        YOLOv13-S~\cite{lei2025yolov13} & 600 & 9 & 21 & 2.98  & 48.0 & 65.2 & 52.0 & - & - & - \\
        RT-DETR-R18~\cite{zhao2024detrs} & 120 & 20 & 60 & 4.61  & 46.5 & 63.8 & 50.4 & 28.4 & 49.8 & 63.0 \\
        RT-DETRv2-S~\cite{lv2024rt} & 120 & 20 & 60 & 4.61  & 48.1 & 65.1 & 52.1$^*$ & 30.2$^*$ & 51.2$^*$ & 64.2$^*$ \\
        RT-DETRv3-R18~\cite{wang2025rt} & 120 & 20 & 60 & 4.61 & 48.1 & 65.6 & 52.0$^*$ & 30.2$^*$ & 51.5$^*$ & 63.9$^*$ \\
        D-FINE-S~\cite{peng2025d} & 120 & 10 & 25 & 3.66 & 48.5 & 65.6 & 52.6 & 29.1 & 52.2 & 65.4 \\
        DEIM-S~\cite{huang2025deim} & 120 & 10 & 25 & 3.66 & \underline{49.0} & \underline{65.9} & \underline{53.1} & \underline{30.4} & 52.6 & \underline{65.7} \\
        \rowcolor{aliceblue} \textbf{{\ourmethod}-S (ours)} & 120 & 10 & 25 & 3.66 & \textbf{49.8} & \textbf{67.1} & \textbf{54.0} & \textbf{30.6} & \textbf{53.4} & \textbf{67.3} \\

        \midrule
        YOLOv10-M~\cite{wang2024yolov10} & 500 & 15 & 59 & 4.74 & 51.1 & 68.1 & 55.8 & 33.8 & 56.5 & 67.0 \\
        YOLO11-M~\cite{yolo11} & 500 & 20 & 68 & 4.85 & 51.5 & 68.1$^*$ & 55.8$^*$ & - & - & - \\
        YOLOv12-M~\cite{tian2025yolov12} & 600 & 20 & 68 & 4.96 & 52.5 & 69.6 & 57.1 & \textbf{35.7} & \textbf{58.2} & 68.8 \\
        RT-DETR-R34~\cite{zhao2024detrs} & 120 & 31 & 92 & 6.91 & 48.9 & 66.8 & 52.9 & 30.6 & 52.4 & 66.3 \\
        RT-DETRv2-M~\cite{lv2024rt} & 120 & 31 & 92 & 6.91 & 49.9 & 67.5 & 54.1$^*$ & 32.0$^*$ & 53.2$^*$ & 66.5$^*$ \\
        RT-DETRv3-R34~\cite{wang2025rt} & 120 & 31 & 92 & 6.91 & 49.9 & 67.7 & 53.9$^*$ & 31.7$^*$ & 54.0$^*$ & 66.2$^*$ \\
        D-FINE-M~\cite{peng2025d} & 120 & 19 & 57 & 5.91 & 52.3 & 69.8 & 56.4 & 33.2 & 56.5 & \underline{70.2} \\
        DEIM-M~\cite{huang2025deim} & 90 & 19 & 57 & 5.91 & \underline{52.7} & \underline{70.0} & \underline{57.3} & 35.3 & 56.7 & 69.5 \\
        % DEIMv2-S~\cite{huang2025real} & 120 & 10 & 26 & 5.78 & 50.9 & 68.3 & 55.1 & 31.4 & 55.3 & 70.3 \\
        \rowcolor{aliceblue} \textbf{{\ourmethod}-M (ours)} & 90 & 19 & 57 & 5.91 & \textbf{53.7} & \textbf{71.0} & \textbf{58.4} & \underline{35.6} & \underline{57.4} & \textbf{71.2} \\

        \midrule
        YOLOv10-L~\cite{wang2024yolov10} & 500 & 24 & 120 & 7.38 & 53.2 & 70.1 & 58.1 & 35.8 & 58.5 & 69.4 \\
        YOLO11-L~\cite{yolo11} & 500 & 25 & 87 & 6.33 & 53.4 & 69.7$^*$ & 58.3$^*$ & - & - & - \\
        YOLOv12-L~\cite{tian2025yolov12} & 600 & 27 & 89 & 6.85 & 53.7 & 70.7 & 58.5 & \underline{36.9} & 59.5 & 69.9 \\
        YOLOv13-L~\cite{lei2025yolov13} & 600 & 28 & 88 & 8.63  & 53.4 & 70.9 & 58.1 & - & - & - \\
        RT-DETR-R50~\cite{zhao2024detrs} & 72 & 42 & 136 & 9.29 & 53.1 & 71.3 & 57.7 & 34.8 & 58.0 & 70.0 \\
        RT-DETRv2-L~\cite{lv2024rt} & 72 & 42 & 136 & 9.29 & 53.4 & 71.6 & 57.4$^*$ & 36.1$^*$ & 57.9$^*$ & 70.8$^*$ \\
        RT-DETRv3-R50~\cite{wang2025rt} & 120 & 42 & 136 & 9.29 & 53.4 & 71.7 & 57.3$^*$ & 35.4$^*$ & 57.4$^*$ & 69.8$^*$ \\
        D-FINE-L~\cite{peng2025d} & 72 & 31 & 91 & 8.07 & 54.0 & 71.6 & 58.4 & 36.5 & 58.0 & \underline{71.9} \\
        DEIM-L~\cite{huang2025deim} & 50 & 31 & 91 & 8.07 & \underline{54.7} & \underline{72.4} & \underline{59.4} & \underline{36.9} & \underline{59.6} & 71.8 \\
        % DEIMv2-M~\cite{huang2025real} & 90 & 18 & 52 & 8.80 & 53.0 & 70.2 & 57.6 & 34.2 & 57.4 & 71.5 \\
        \rowcolor{aliceblue} \textbf{{\ourmethod}-L (ours)} & 50 & 31 & 91 & 8.07 & \textbf{55.4} & \textbf{73.0} & \textbf{60.3} & \textbf{37.1} & \textbf{60.1} & \textbf{72.9} \\

        \midrule
        YOLOv10-X~\cite{wang2024yolov10} & 500 & 30 & 160 & 10.45 & 54.4 & 71.3 & 59.3 & 37.0 & 59.8 & 70.9 \\
        YOLO11-X~\cite{yolo11} & 500 & 57 & 195 & 11.50 & 54.7 & 71.3$^*$ & 59.7$^*$ & - & - & - \\
        YOLOv12-X~\cite{tian2025yolov12} & 600 & 59 & 199 & 11.80 & 55.2 & 72.0 & 60.2 & \textbf{39.6} & 60.7 & 70.9 \\
        YOLOv13-X~\cite{lei2025yolov13} & 600 & 64 & 199 & 14.67  & 54.8 & 72.0 & 59.8 & - & - & - \\
        RT-DETR-R101~\cite{zhao2024detrs} & 72 & 76 & 259 & 13.88 & 54.3 & 72.7 & 58.6 & 36.0 & 58.8 & 72.1 \\
        RT-DETRv2-X~\cite{lv2024rt} & 72 & 76 & 259 & 13.88 & 54.3 & 72.8 & 58.8$^*$ & 35.8$^*$ & 58.8$^*$ & 72.1$^*$ \\
        RT-DETRv3-R101~\cite{wang2025rt} & 120 & 76 & 259 & 13.88 & 54.6 & 73.1 & - & - & - & - \\
        D-FINE-X~\cite{peng2025d} & 72 & 62 & 202 & 12.90 & 55.8 & 73.7 & 60.2 & 37.3 & 60.5 & 73.4 \\
        DEIM-X~\cite{huang2025deim} & 50 & 62 & 202 & 12.90 & \underline{56.5} & \underline{74.0} & \underline{61.5} & 38.8 & \underline{61.4} & \underline{74.2} \\
        % DEIMv2-L~\cite{huang2025real} & 60 & 32 & 96 & 10.47 & 56.0 & 73.4 & 60.9 & 37.5 & 60.8 & 75.2 \\
        \rowcolor{aliceblue} \textbf{{\ourmethod}-X (ours)} & 50 & 62 & 202 & 12.90 & \textbf{57.0} & \textbf{74.6} & \textbf{62.1} & \underline{39.5} & \textbf{61.9} & \textbf{74.8} \\
        \bottomrule
    \end{tabular}
    \end{adjustbox}
    \end{center}
    \vspace{-5mm}
\end{table}

\subsection{Ablation study}
\label{subsec:ablation}

We conduct a series of ablation experiments to verify the effectiveness of the proposed modules. Unless stated otherwise, all ablation experiments are trained for \textbf{36 epochs}. Unspecified hyperparameters or configurations follow the best settings for the corresponding experimental setup.

\begin{figure*}[t]
    \centering
    \includegraphics[width=1\linewidth]{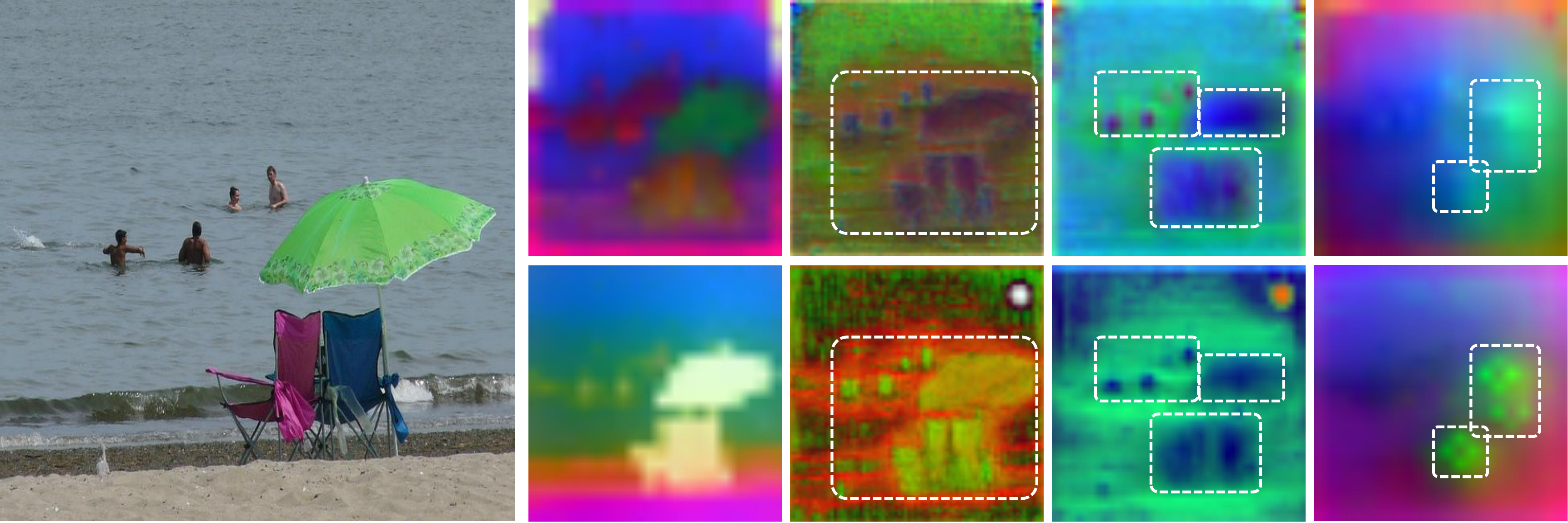}
    \caption{\textbf{Comparison of dense features.} We compare the feature map quality of DEIM-L (top) and \ourmethod-L (bottom) by projecting dense outputs to RGB space using PCA. The visualization reveals that our DSI module substantially enhances the semantic representation of deep features. From left to right: input image, AIFI feature map $F_5$, and CCFF features $P_3$, $P_4$, $P_5$.}
    \label{fig:feature_map_comparison}
\end{figure*}

\noindent \textbf{Feature visualization.}
\cref{fig:feature_map_comparison} visually compares the feature maps between DEIM-L and our RT-DETRv4-L. Notably, our model enriches the semantic content of the AIFI feature $F_5$, leading to more precise and distinguishable object contours and backgrounds across the subsequent multi-scale features $P_3$, $P_4$, and $P_5$. In particular, $P_5$ exhibits a markedly stronger and more concentrated response to object regions.

\begin{table}[!tb]
\centering
\scriptsize
\caption{\textbf{Results of ablation on DSI and GAM across multiple detectors.} Our method brings consistent and significant gains with zero additional inference cost.}
\label{tab:ablation_main}
\vspace{-2mm}
\small
\begin{tabularx}{\textwidth}{>{\raggedright\arraybackslash}p{3.8cm} | C C C}
    \toprule
    \textbf{Method} & \textbf{AP$^{val}$} & \textbf{AP$^{val}_{50}$} & \textbf{AP$^{val}_{75}$} \\
    \midrule
    \textbf{RT-DETRv2-L} & 52.1 & 70.2 & 56.7 \\
    w/ DSI & 52.3 \textcolor{cvprblue}{(+0.2)} & 70.4 \textcolor{cvprblue}{(+0.2)} & 56.4 \textcolor{red}{(-0.3)} \\
    w/ DSI+GAM & \textbf{52.6} \textbf{\textcolor{cvprblue}{(+0.5)}} & \textbf{70.7} \textbf{\textcolor{cvprblue}{(+0.5)}} & \textbf{56.8} \textbf{\textcolor{cvprblue}{(+0.1)}} \\
    \midrule
    \textbf{D-FINE-L} & 53.1 & 70.8 & 57.4 \\
    w/ DSI & 53.2 \textcolor{cvprblue}{(+0.1)} & 70.8 \textcolor{cvprblue}{(+0)} & 57.7 \textcolor{cvprblue}{(+0.3)} \\
    w/ DSI+GAM & \textbf{53.4} \textbf{\textcolor{cvprblue}{(+0.3)}} & \textbf{71.1} \textbf{\textcolor{cvprblue}{(+0.3)}} & \textbf{58.0} \textbf{\textcolor{cvprblue}{(+0.6)}} \\
    \midrule
    \textbf{DEIM-L} & 53.8 & 71.4 & 58.5 \\
    w/ DSI & 53.9 \textcolor{cvprblue}{(+0.1)} & 71.3 \textcolor{red}{(-0.1)} & 58.8 \textcolor{cvprblue}{(+0.3)} \\
    w/ DSI+GAM & \textbf{54.3} \textbf{\textcolor{cvprblue}{(+0.5)}} & \textbf{71.8} \textbf{\textcolor{cvprblue}{(+0.4)}} & \textbf{59.0} \textbf{\textcolor{cvprblue}{(+0.5)}} \\
    \bottomrule
\end{tabularx}
\end{table}
\begin{table}[!h]
\centering
\caption{\textbf{Results of ablation on semantic injection position.} We compare the effectiveness of applying DSI at different positions of \textbf{DEIM-L}, corresponding to the strategies in~\Cref{fig:module_design}. Aligning only the AIFI output (\(F_5\)) yields the best performance.}
\label{tab:semantic_location}
\vspace{-2mm}
\small
\begin{tabularx}{\textwidth}{>{\raggedright\arraybackslash}p{2cm} | C C C C | C C C}
    \toprule
    \textbf{Position} & \textbf{$S_{3}$} & \textbf{$S_{4}$} & \textbf{$S_{5}$} & \textbf{$F_{5}$} & \textbf{AP$^{val}$} & \textbf{AP$^{val}_{50}$} & \textbf{AP$^{val}_{75}$} \\
    \midrule
    Baseline & - & - & - & - & 53.8 & 71.4 & 58.5 \\
    \midrule
    \multirow{4}{*}{Backbone} & \checkmark & - & - & - & 53.7 & 71.2 & 58.4 \\
    & - & \checkmark & - & - & 53.7 & 71.3 & 58.4 \\
    & - & - & \checkmark & - & 53.8 & 71.3 & 58.5 \\
    & \checkmark & \checkmark & \checkmark & - & 53.7 & 71.3 & 58.5 \\
    \midrule
    Hybrid & \checkmark & \checkmark & \checkmark & \checkmark & 53.8 & 71.4 & 58.4 \\
    \midrule
    AIFI & - & - & - & \checkmark & \textbf{54.3} & \textbf{71.8} & \textbf{59.0} \\
    \bottomrule
\end{tabularx}
\end{table}

\noindent \textbf{Ablation on DSI and GAM.}
We first assess the effectiveness of DSI and GAM.
As shown in Tab.~\ref{tab:ablation_main}, applying DSI can bring a slight performance improvement, while further applying GAM can significantly improve the performance gain (0.5 AP), which fully proves the effectiveness of both.
To verify the general applicability of our method, we also conduct experiments on RT-DETRv2 and D-FINE, and the results show that our method can bring consistent performance gains to different detectors.

\begin{table}[!t]
\centering
\small
\caption{\textbf{Results of ablation on the feature projector.}}
\label{tab:ablation_projector}
\vspace{-3mm}
\begin{tabularx}{\textwidth}{>{\raggedright\arraybackslash}p{3.8cm}  | C C C}
    \toprule
    \textbf{Projector Arch.\hspace{1cm}} & \textbf{AP$^{val}$} & \textbf{AP$^{val}_{50}$} & \textbf{AP$^{val}_{75}$} \\
    \midrule
    \textbf{DEIM-L} & 53.8 & 71.4 & 58.5 \\
    \midrule
    w/ 1x1 Conv & 53.8 & 71.5 & 58.5 \\
    w/ MLP & 54.2 & 71.7 & 59.0\\
    w/ Linear & \textbf{54.3} \textcolor{cvprblue}{(+0.5)} & \textbf{71.8} \textcolor{cvprblue}{(+0.4)} & \textbf{59.0} \textcolor{cvprblue}{(+0.5)} \\
    \bottomrule
\end{tabularx}
\end{table}
\begin{table}[!t]
\centering
\caption{\textbf{Results of ablation on the semantic teacher.}}
\label{tab:semantic_teacher}
\vspace{-3mm}
\begin{tabularx}{\textwidth}{>{\raggedright\arraybackslash}p{3.8cm}  | C C C}
    \toprule
    \textbf{Semantic Teacher} & \textbf{AP$^{val}$} & \textbf{AP$^{val}_{50}$} & \textbf{AP$^{val}_{75}$} \\
    \midrule
    \textbf{DEIM-L} & 53.8 & 71.4 & 58.5 \\
    \midrule
    w/ MAE-ViT-B & 53.9 & 71.4 & 58.6 \\
    w/ DINOv2-ViT-Base & 54.1 & 71.6 & 58.8\\
    w/ DINOv3-ViT-Base & \textbf{54.3} \textcolor{cvprblue}{(+0.5)} & \textbf{71.9} \textcolor{cvprblue}{(+0.5)} & \textbf{59.0} \textcolor{cvprblue}{(+0.5)} \\
    \bottomrule
\end{tabularx}
\end{table}
\begin{table}[!t]
\centering
\caption{\textbf{Results of ablation on the alignment loss.} The cosine similarity loss demonstrates superior performance. DEIM-M is adopted as the baseline, with all models trained for 90 epochs and 12 EMA epochs.}
\label{tab:ablation_loss}
\vspace{-3mm}
\small
\begin{tabularx}{\textwidth}{>{\raggedright\arraybackslash}p{3.8cm}  | C C C}
    \toprule
    \textbf{Loss Function} & \textbf{AP$^{val}$} & \textbf{AP$^{val}_{50}$} & \textbf{AP$^{val}_{75}$} \\
    \midrule
    \textbf{DEIM-M} & 52.5 & 69.9 & 57.2 \\
    \midrule
    w/ MSE Loss & 52.7 & 70.0 & 57.4 \\
    w/ Cosine Similarity & \textbf{53.7} \textcolor{cvprblue}{(+1.2)} & \textbf{71.0} \textcolor{cvprblue}{(+1.1)} & \textbf{58.4} \textcolor{cvprblue}{(+1.2)} \\
    \bottomrule
\end{tabularx}
\end{table}
\begin{table}[!t]
\centering
\setlength{\tabcolsep}{4.3pt}
\renewcommand{\arraystretch}{1}
\caption{\textbf{Ablation on the loss weighting strategy.} GAM consistently outperforms the static weighting. DEIM-L is adopted as the baseline, with all models trained for 50 epochs and 8 EMA epochs.}
\label{tab:ablation_gam}
\vspace{-6mm}
\resizebox{\textwidth}{!}{
\begin{tabular}{c|cccccccccccc}
    \multicolumn{7}{l}{}\\
    \toprule
    \textbf{$\lambda$} & 0.1 & 0.2 & 0.4 & 1 & 2 & \underline{4} & \underline{10} & \textbf{20} & \underline{30} & 50 & 100 & \textbf{GAM}\\
    \arrayrulecolor{gray}\midrule
    \textbf{AP$^{val}$} & 54.7 & 54.7 & 54.8 & 54.9 & 54.9 & \underline{55.0} & \underline{55.0} & \textbf{55.1} & \underline{55.0} & 54.9 & 54.6 &  \textbf{55.4}  \\
    \arrayrulecolor{black}\bottomrule
\end{tabular}
}
\end{table}

\begin{figure}[!t]
    \centering
    \vspace{2mm}
    \includegraphics[width=0.6\linewidth]{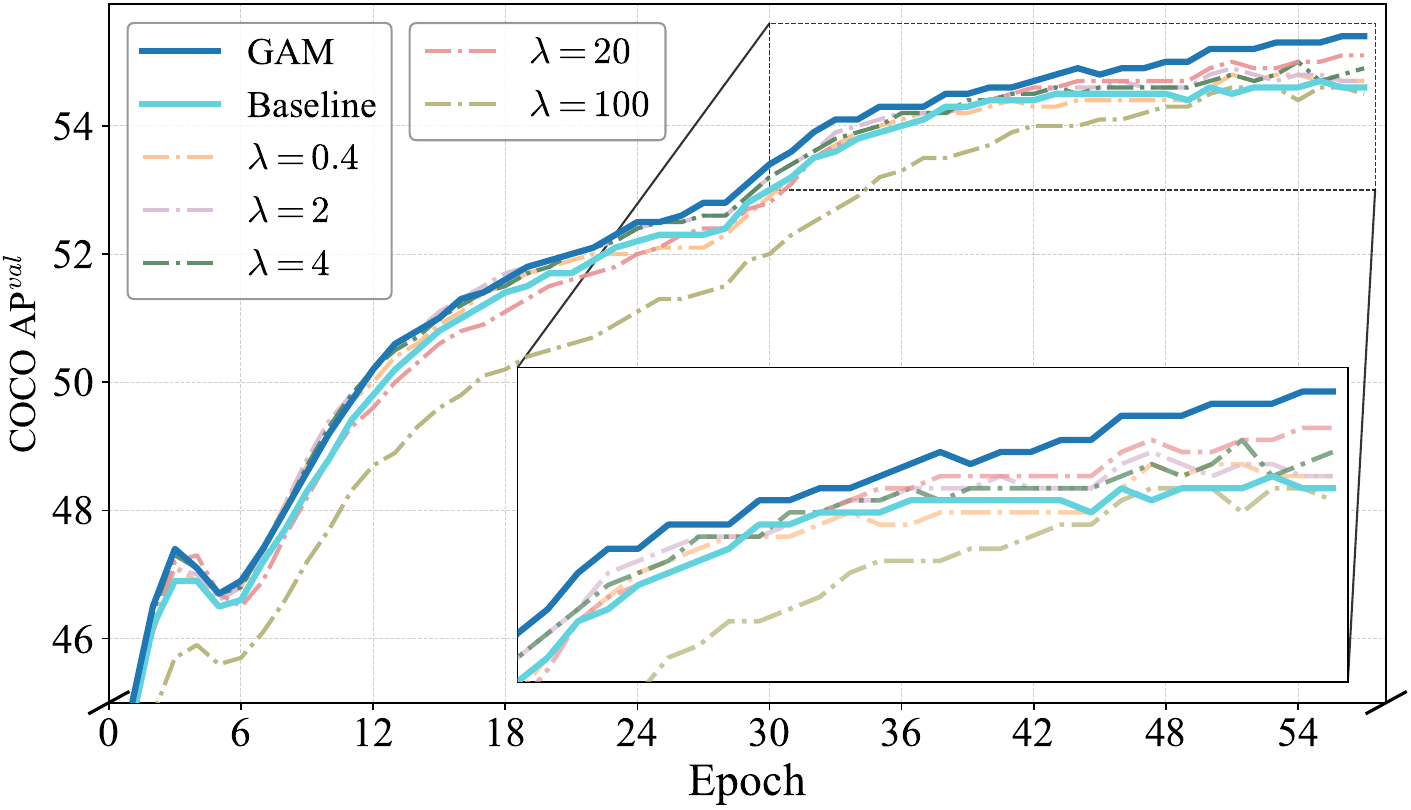}
    \caption{\textbf{AP$^{val}$ evolution on COCO during training.}
    }
    \label{fig:gam_training_curve}
    \vspace{-5mm}
\end{figure}

\noindent \textbf{Ablation on semantic injection position.}
To validate the choice of injection position, we compare the strategies shown in~\cref{fig:module_design}. Results in~\cref{tab:semantic_location} indicate that directly applying semantic supervision to backbone features (\(S_3\), \(S_4\), or \(S_5\)) individually or jointly yields no improvement. Similarly, the hybrid approach (strategy (b)) that aligns both backbone and \(F_5\) features provides no gain. 

In contrast, our design (strategy (c)), which aligns only the AIFI output \(F_5\), achieves a clear 0.5 AP improvement (54.3 AP). 
This demonstrates that maintaining richer semantics in \(F_5\) is crucial for enhancing detection performance, as it plays a key role in propagating high-level semantic information to subsequent feature hierarchies. Furthermore, the ineffectiveness of the hybrid approach suggests that simultaneously aligning features from the CNN-based backbone and the Transformer-based AIFI may introduce optimization conflicts or semantic misalignment. Our chosen strategy is not only more effective but also more efficient. It avoids the complexity of multiple intermediate projections and interpolations, and the gradient from the single \(F_5\) alignment loss naturally flows back to synergistically update both AIFI and the backbone, achieving consistent enhancement with a single, targeted objective.

\noindent \textbf{Design of the feature projector.}
The feature projector is a crucial component for bridging the student and teacher feature spaces. We explore several architectural choices, as detailed in~\cref{tab:ablation_projector}. A linear-based projector yields the best results, striking an optimal balance between expressive power and parameter efficiency.

\noindent \textbf{Choice of semantic teacher.}
\cref{tab:semantic_teacher} shows that DINOv3-ViT-Base outperforms DINOv2-ViT-Base and MAE-ViT-B as the semantic teacher, 
confirming that models with stronger semantic priors acquired from large-scale self-supervised learning serve better in this role.  
Moreover, our method remains agnostic to the type of VFMs, allowing detectors to continuously benefit from future advances in pretrained VFMs, providing a flexible and training-friendly solution.

\noindent \textbf{Ablation on loss function.}
We further study the alignment loss of DSI.
We compare the Mean Squared Error (MSE) loss and the cosine similarity loss in~\cref{tab:ablation_loss}, the cosine similarity loss surpasses MSE.

\noindent \textbf{Comparison of GAM and static weights.}
Finally, we validate the proposed GAM against static weights. The superiority of GAM in navigating the training dynamics is illustrated in~\cref{fig:gam_training_curve}, showing AP$^{val}$ over training epochs. As shown, GAM consistently surpasses the baseline and all static weight configurations, demonstrating a clear and stable performance advantage throughout training.

\cref{tab:ablation_gam} details the results for static weight and GAM. For static weighting, performance peaks at $\lambda=20$, but degrades with either smaller or larger values. However, observing the training curve in~\cref{fig:gam_training_curve}, we find that even this optimal static weight ($\lambda=20$) leads to slower convergence in the early-to-mid stages, highlighting the inherent limitations of a fixed hyperparameter. Our experiments indicate that GAM achieves the best performance (55.4 AP).
\section{Discussion}

To further highlight the advantages of our framework over existing methods~\cite{tian2025yolov12,lei2025yolov13,peng2025d,huang2025deim}, we discuss it from three perspectives: deployment efficiency, scalability, and training efficiency.

\noindent \textbf{Deployment Efficiency.}
Our method introduces zero modification to detector architectures and does not alter the inference pipeline, ensuring that no additional computational cost or latency is incurred. This deployment-friendly property is crucial for industrial applications, where real-time detectors are tightly integrated into existing systems and hardware-constrained environments.

\noindent \textbf{Scalability.}
The framework is general and can be applied seamlessly to detectors with diverse architectures, including CNN-based and transformer-based detectors. It enables all types of real-time detectors to quickly benefit from the rapid progress of Vision Foundation Models (VFMs). 
Moreover, the framework is not restricted to any specific type or scale of VFM.
It can flexibly incorporate different foundation models, such as DINOv3~\cite{simeoni2025dinov3}, MAE~\cite{he2022masked}, or CLIP~\cite{radford2021learning}, and even benefit from arbitrarily large models for distilling semantics into real-time detectors.
This flexibility also opens up promising directions for multi-VFM semantic integration, further demonstrating the framework's generality and scalability.

\noindent \textbf{Training Efficiency.}
Our approach is lightweight and easy to implement. Since neither the detector structure nor the optimization pipeline is modified, the additional training cost remains minimal. 
Meanwhile, as VFMs continue to advance, the semantic teacher can be updated with stronger pretrained models while maintaining the same training cost. 
This efficiency highlights the practicality of our method for large-scale applications and real-time industrial deployment.
\section{Conclusion}

In this work, we present a cost-effective and adaptable semantic distillation framework that enhances real-time DETR-based object detectors without increasing inference or deployment overhead.
Through the proposed Deep Semantic Injector (DSI) and Gradient-guided Adaptive Modulation (GAM), our method effectively transfers high-level semantics from Vision Foundation Models to lightweight detectors in a stable and task-aligned manner.
Extensive experiments on COCO demonstrate consistent and significant performance gains across multiple model scales, culminating in the state-of-the-art RT-DETRv4 series.
These results highlight the effectiveness of explicit semantic supervision in bridging the gap between large-scale foundation models and resource-efficient detection architectures.
Overall, our work provides a practical pathway toward unlocking the potential of foundation models for efficient visual perception.

\section*{Acknowledgments}
This work was supported in part by the New Generation Artificial Intelligence-National Science and Technology Major Project (No. 2025ZD0122702), the Shenzhen Medical Research Funds in China (No. B2302037), Natural Science Foundation of China (No. 61972217, 32071459, 62176249, 62006133, 62271465, 62406167), and AI for Science (AI4S)-Preferred Program, Peking University Shenzhen Graduate School, China, and the Shenzhen Loop Area Institute under grant FPF10120250014.

\bibliographystyle{splncs04}
\bibliography{main}

\clearpage
\maketitlesupplementary

\renewcommand{\thefootnote}{\fnsymbol{footnote}}

\renewcommand{\thetable}{\Alph{table}}
\renewcommand{\thesection}{\Alph{section}}
\renewcommand{\theequation}{\Alph{equation}}
\renewcommand{\thefigure}{\Alph{figure}}

\setcounter{table}{0}
\setcounter{section}{0}
\setcounter{figure}{0}
\setcounter{equation}{0}

\definecolor{darkgreen}{HTML}{539165}
\newcommand{\increase}[1]{{
  \fontsize{9.5pt}{0.5em}\selectfont({\color{darkgreen}{$\uparrow$~\textbf{#1}}})
}}

%%%%%%%%%%%%%%%%%%%%%%%%%%%%%%%%%%%%%%%%%%%%%%%%%%%%%%%%%%%%%%%%%%%%%%
\begin{figure}[!h]
\centering
\includegraphics[width=\linewidth]{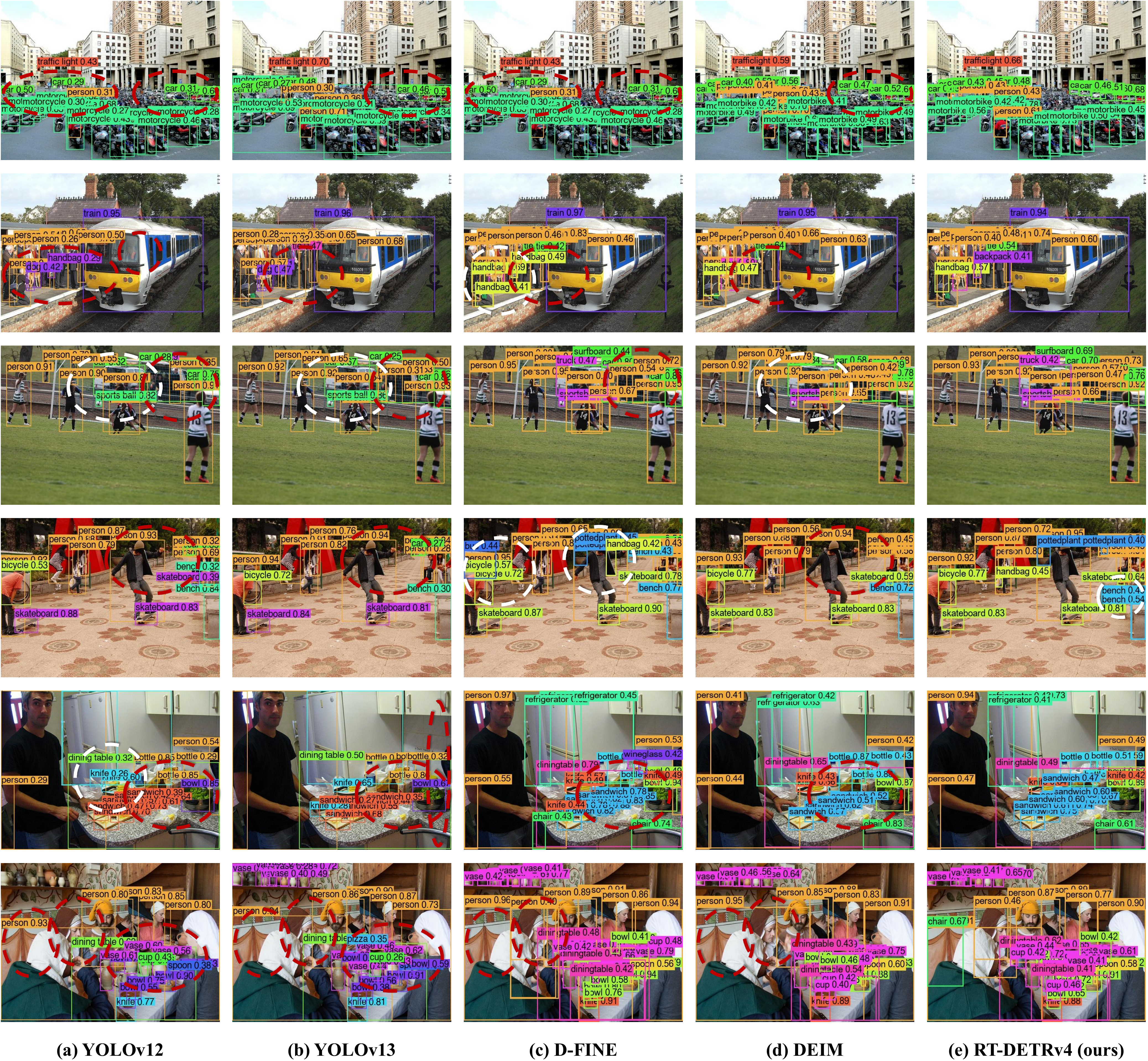}
\caption{\textbf{Visualization of RT-DETRv4-L predictions.} 
Comparison with novel real-time detectors including YOLOv12~\cite{tian2025yolov12}, YOLOv13~\cite{lei2025yolov13}, D-FINE~\cite{peng2025d}, and DEIM~\cite{huang2025deim}. Red circles indicate missed detections, while white circles denote duplicate detections or misclassifications. }

\label{fig:vis_1}
\vspace*{-2mm}
\end{figure}
%%%%%%%%%%%%%%%%%%%%%%%%%%%%%%%%%%%%%%%%%%%%%%%%%%%%%%%%%%%%%%%%%%%%%%

%%%%%%%%%%%%%%%%%%%%%%%%%%%%%%%%%%%%%%%%%%%%%%%%%%%%%%%%%%%%%
\section{Experimental setup}
\label{sec:exp_setup}
%%%%%%%%%%%%%%%%%%%%%%%%%%%%%%%%%%%%%%%%%%%%%%%%%%%%%%%%%%%%%

\noindent \textbf{Dataset and metric.} 
All experiments are conducted on the COCO 2017~\cite{lin2014microsoft} dataset, using the \texttt{train2017} split for training and \texttt{val2017} for evaluation. We report the standard COCO metrics, including AP~(averaged over uniformly sampled IoU thresholds ranging from 0.50 to 0.95 with a step size of 0.05), AP$_{50}$, AP$_{75}$, as well as AP at different scales: AP$_S$, AP$_M$, AP$_L$.

\noindent \textbf{Implementation details.}
Our experiments are built upon the RT-DETR architecture~\cite{zhao2024detrs}, incorporating architectural and training refinements from RT-DETRv2~\cite{lv2024rt}, D-FINE~\cite{peng2025d}, and DEIM~\cite{huang2025deim}. For fair comparison, the core hyperparameters are kept consistent with those of the corresponding baseline models. 
The DSI module employs a pre-trained and frozen DINOv3-ViT-Base model as the semantic teacher. All evaluations are conducted using the COCO AP metrics, and inference latency (ms) is measured on a single NVIDIA T4 GPU under TensorRT FP16 precision.
The hyperparameters of RT-DETRv4 that differ from those of the corresponding-size baselines (RT-DETRv2~\cite{lv2024rt}, D-FINE~\cite{peng2025d}, and DEIM~\cite{huang2025deim}) are summarized in~\Cref{tab:hyperparams}.

%%%%%%%%%%%%%%%%%%%%%%%%%%%%%%%%%%%%%%%%%%%%%%%%%%%%%%%%%%%%%%%%%%%%%%
\begin{figure}[!t]
\centering
\includegraphics[width=\linewidth]{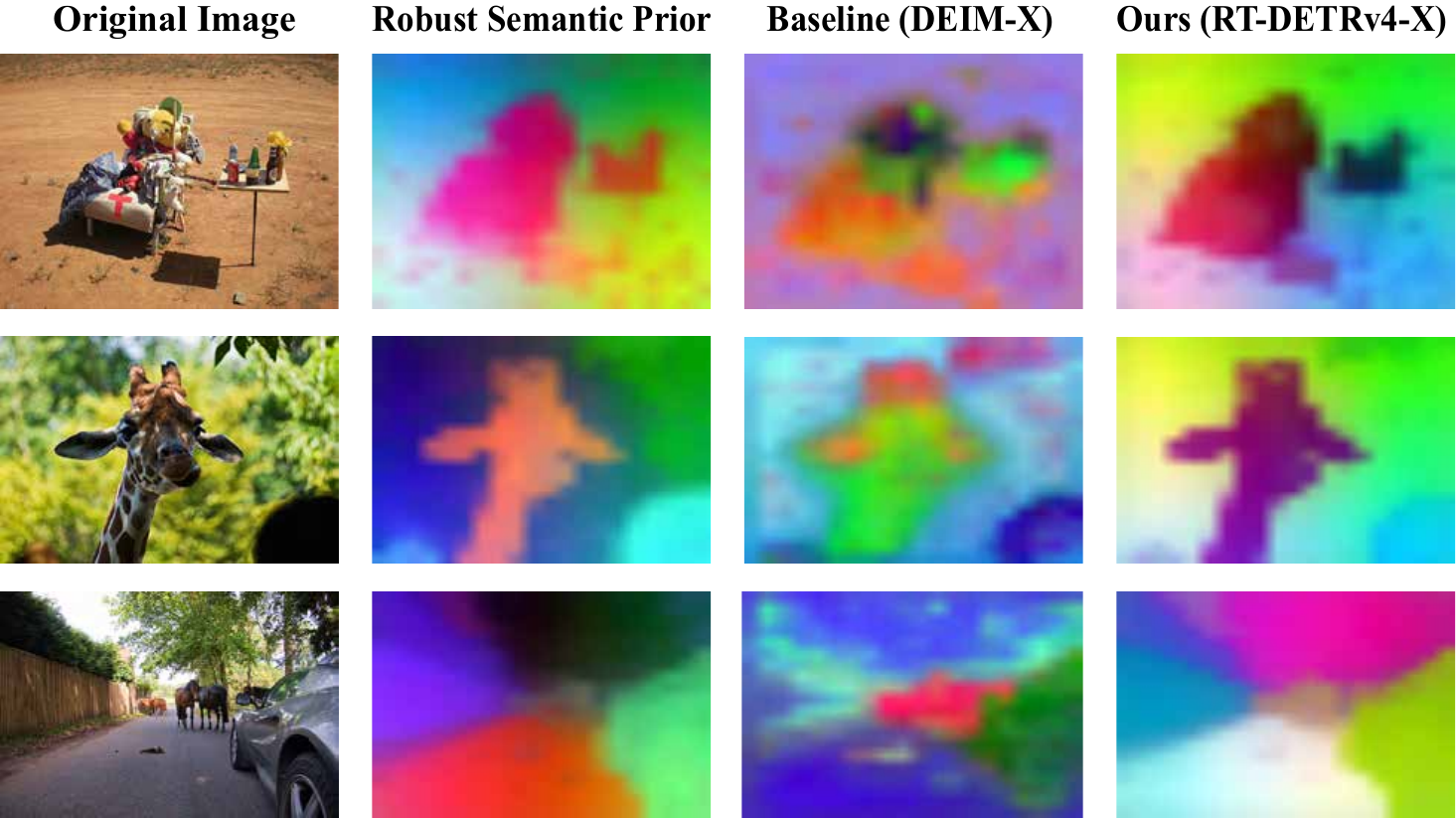}
\caption{\textbf{Visualization of the deepest feature $F_5$ using PCA.} 
We compare the semantic representations of the deepest feature $F_5$ between the baseline DEIM~\cite{huang2025deim} (without semantic injection) and our method (trained with semantic injection but without retaining semantic teacher at inference). }
\label{fig:vis_2}
\vspace*{-2mm}
\end{figure}
%%%%%%%%%%%%%%%%%%%%%%%%%%%%%%%%%%%%%%%%%%%%%%%%%%%%%%%%%%%%%%%%%%%%%%

\begin{table}[!htbp] 
    \centering

    \begin{minipage}{\linewidth}
    \centering
    \caption{\textbf{Detailed hyper-parameters of RT-DETRv4.} 
    The training hyper-parameters for RT-DETRv4 models, focusing on the settings that diverge from the baselines (RT-DETRv2~\cite{lv2024rt}, D-FINE~\cite{peng2025d}, and DEIM~\cite{huang2025deim}).
    }
    \label{tab:hyperparams}
        \begin{adjustbox}{width=\textwidth}
        \begin{tabular}{l|cccc} 
            \toprule
            \textbf{Hyper-parameters} & \textbf{S} & \textbf{M} & \textbf{L} & \textbf{X}\\
            \midrule
            Semantic Teacher & DINOv3-ViT-Base & DINOv3-ViT-Base & DINOv3-ViT-Base & DINOv3-ViT-Base \\
            \midrule
            Patch Size & 16 & 16 & 16 & 16 \\
            Projector Architecture & Linear & Linear & Linear & Linear \\
            Projector Dimension & 768 & 768 & 768 & 768 \\
            Normalization Mean & [0.485, 0.456, 0.406] & [0.485, 0.456, 0.406] & [0.485, 0.456, 0.406] & [0.485, 0.456, 0.406] \\
            Normalization Std & [0.229, 0.224, 0.225] & [0.229, 0.224, 0.225] & [0.229, 0.224, 0.225] & [0.229, 0.224, 0.225] \\
            \midrule
            Training Epochs & 120 & 90 & 50 & 50 \\
            EMA Epochs & 12 & 12 & 8 & 8 \\
            \midrule
            GAM & \checkmark & \checkmark & \checkmark & \checkmark \\
            \midrule
            GAM Warmup Weight & 1 & 1 & 3 & 2\\
            GAM Rho (\%) & 11 & 3.5 & 2 & 2\\
            GAM Delta (\%) & 1 & 0.25 & 0.1 & 0.25 \\
            GAM EMA Weight & 4 & 3 & 3 & 4 \\
            \bottomrule
        \end{tabular}
        \end{adjustbox}
    \end{minipage}

    \vspace{8mm}

    \begin{minipage}{\linewidth}
        \centering
        \caption{
        \textbf{Training cost comparison across model scales.}
        Total GPU hours denote the total training time measured on NVIDIA GeForce RTX 3090 GPUs.
        Memory Per Image reports the average GPU memory usage per image during training.
        }
        \label{tab:training_cost}
        \scriptsize
        \setlength{\tabcolsep}{6pt}
        \begin{adjustbox}{width=\textwidth}
        \begin{tabular}{l|c|ccc|ccc}
        \toprule
        \textbf{Model} & \textbf{Training Epochs} &
        \multicolumn{3}{c|}{\textbf{Total GPU Hours}} &
        \multicolumn{3}{c}{\textbf{Memory Per Image (GiB)}} \\[3pt]
        
        \cmidrule(lr){3-5} \cmidrule(lr){6-8}
        & &
        \textbf{Baseline} & \textbf{Ours} & \textbf{Increase} &
        \textbf{Baseline} & \textbf{Ours} & \textbf{Increase} \\[2pt]
        
        \midrule
        S & 132 (120+12 EMA) & 202.00 & 235.00 & +16.3\% & 0.943 & 0.974 & +3.3\% \\
        M & 102 (90+12 EMA)  & 234.70 & 257.41 & +9.7\%  & 1.084 & 1.119 & +3.2\% \\
        L & 58 (50+8 EMA)    & 152.47 & 166.68 & +9.3\%  & 1.523 & 1.535 & +0.8\% \\
        X & 58 (50+8 EMA)    & 206.67 & 219.48 & +6.2\%  & 1.983 & 2.003 & +1.0\% \\
        \bottomrule
        \end{tabular}
        \end{adjustbox}
    \end{minipage}

    \vspace{8mm}

    \begin{minipage}{\linewidth}
    \centering
    \caption{\textbf{Comparison of Average Gradients for Baseline, DSI, and DSI+GAM.} Comparisons of the average gradients (L1 norm) per epoch for the AIFI module and the overall average gradients for all modules under different configurations: Baseline (DEIM-L), w/ DSI ($\lambda=2$), w/ DSI ($\lambda=20$), and w/ DSI+GAM.
    }
    \label{tab:dynamic_gradients}
    \resizebox{\textwidth}{!}{
    \begin{tabular}{ll|rr|rr|rr|rr} 
    \toprule
    \multirow{2}{*}{Module} & \multirow{2}{*}{Epoch} & \multicolumn{2}{c|}{Baseline} & \multicolumn{2}{c|}{DSI ($\lambda=2$)} & \multicolumn{2}{c|}{DSI ($\lambda=20$)} & \multicolumn{2}{c}{DSI+GAM} \\
    \cmidrule(l){3-4} \cmidrule(l){5-6} \cmidrule(l){7-8} \cmidrule(l){9-10}
     & & Avg Abs Norm & \% & Avg Abs Norm & \% & Avg Abs Norm  & \% & Avg Abs Norm & \% \\
    \midrule
    \multirow{5}{*}{AIFI} & 1-12 & 1.5033 & 0.64\% & 2.7968 & 1.22\% & 6.0068 & 2.55\% & 3.9483 & 1.71\% \\
     & 12-24 & 1.1364 & 0.53\% & 2.5365 & 1.17\% & 4.6953 & 2.13\% & 4.2419 & 1.94\% \\
     & 24-36 & 1.0300 & 0.48\% & 2.2927 & 1.08\% & 3.8318 & 1.81\% & 4.0486 & 1.89\% \\
     & 36-48 & 0.9829 & 0.45\% & 2.2152 & 1.03\% & 3.5430 & 1.67\% & 4.1631 & 1.94\% \\
     & 48-58 & 1.0628 & 0.45\% & 2.5098 & 1.08\% & 4.2570 & 1.85\% & 3.3276 & 1.46\% \\
    \midrule
    % Overall
    Backbone & Overall & 194.7938 & 88.04\% & 190.8166 & 86.84\% & 189.0620 & 85.77\% & 189.9492 & 86.20\% \\
    AIFI & Overall & 1.1459 & 0.51\% & 2.4688 & 1.12\% & 4.4740 & 2.01\% & 3.9672 & 1.80\% \\
    CCFF & Overall & 11.7713 & 5.19\% & 12.5404 & 5.58\% & 13.8424 & 6.16\% & 13.2828 & 5.89\% \\
    Decoder & Overall & 13.9638 & 6.26\% & 14.2981 & 6.46\% & 13.5406 & 6.07\% & 13.6393 & 6.11\% \\
    \midrule
    \textbf{Total} & \textbf{Overall} & \textbf{221.6747} & \textbf{100.00\%} & \textbf{220.1239} & \textbf{100.00\%} & \textbf{220.9190} & \textbf{100.00\%} & \textbf{220.8385} & \textbf{100.00\%} \\
    \bottomrule
    \end{tabular}%
    }
    \end{minipage}

\end{table}

\begin{figure}[!t]
    \centering
    % left
    \begin{minipage}{0.48\textwidth}
        \centering
        \includegraphics[width=\linewidth]{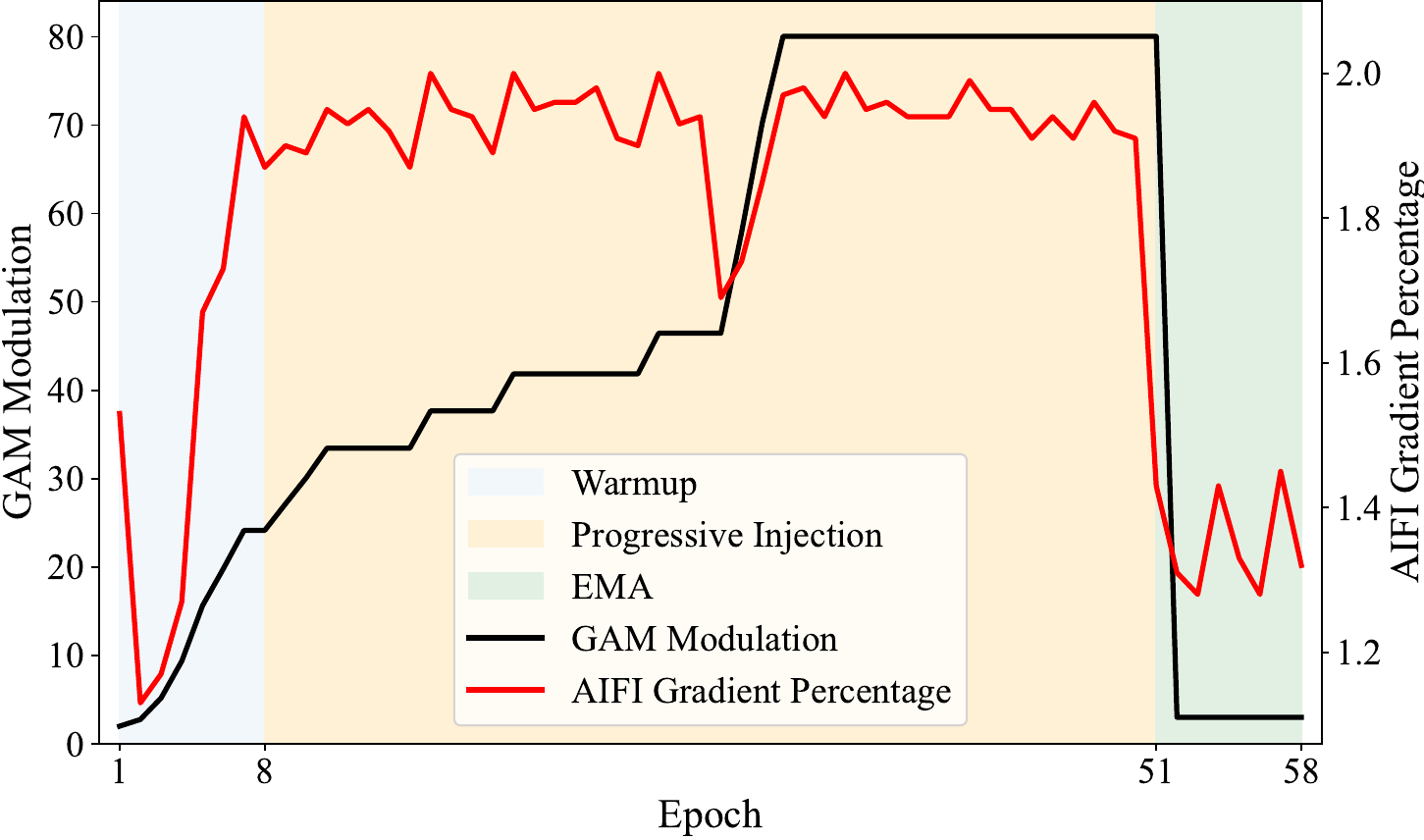}
        \caption{\textbf{Dynamics of GAM During Training.} Dynamic adjustment of the DSI loss weight ($\lambda$) by GAM.}
        \label{fig:gam_dynamics}
    \end{minipage}\hfill
    % right
    \begin{minipage}{0.48\textwidth}
        \centering
        \includegraphics[width=\linewidth]{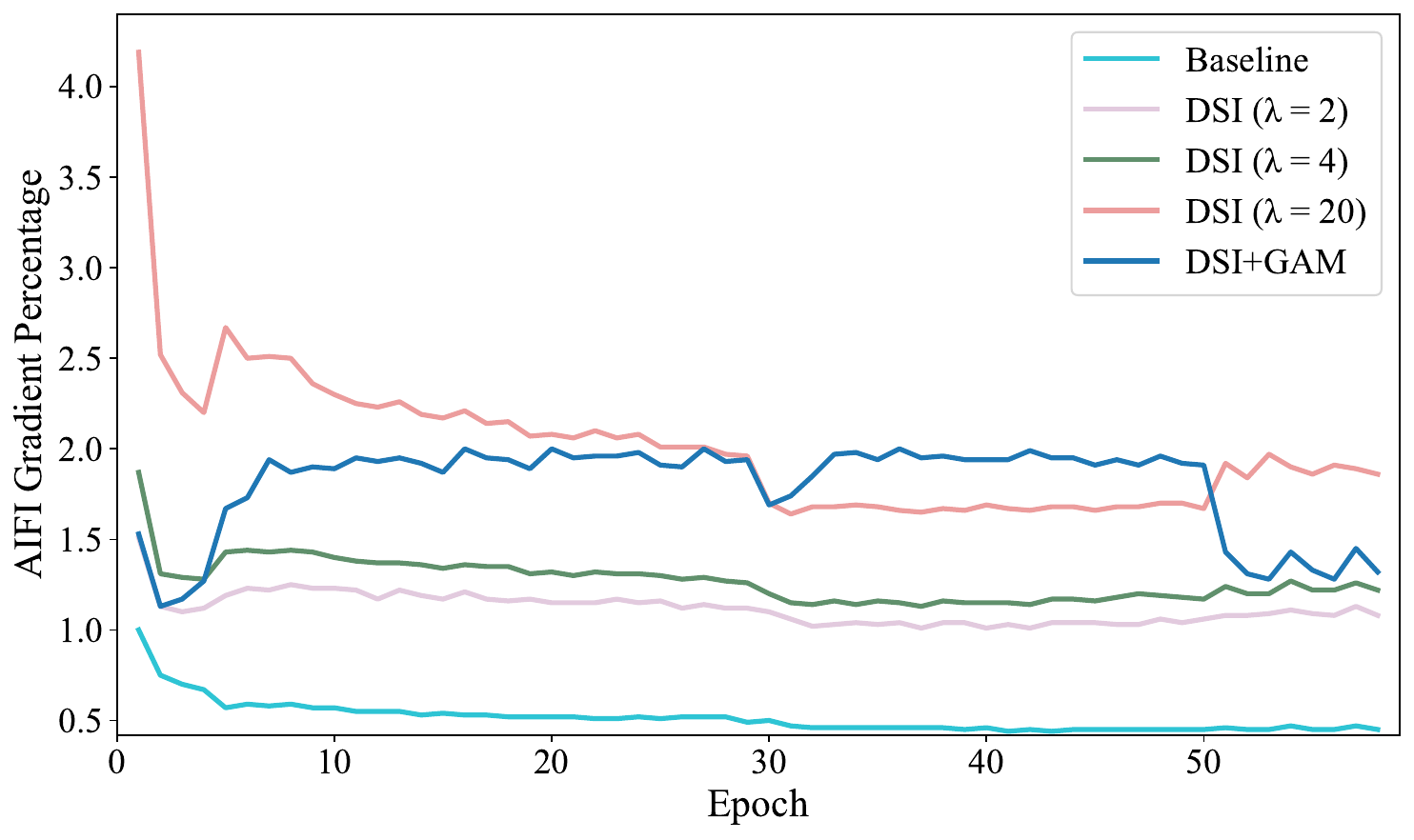}
        \caption{\textbf{Gradient Evolution Over Epochs.} Average gradient trends during training for baseline (DEIM-L), DSI with static $\lambda=2, 4, 20$, and DSI+GAM.}
        \label{fig:gradient_comparison}
    \end{minipage}
    \vspace*{-2mm}
\end{figure}

%%%%%%%%%%%%%%%%%%%%%%%%%%%%%%%%%%%%%%%%%%%%%%%%
\section{Visualization}
\label{sec:visualization}
%%%%%%%%%%%%%%%%%%%%%%%%%%%%%%%%%%%%%%%%%%%%%%%%
We provide qualitative comparisons to further illustrate the effectiveness of RT-DETRv4.  
RT-DETRv4 not only achieves state-of-the-art quantitative results in terms of AP but also demonstrates robust qualitative performance in challenging scenarios. It can accurately detect objects in complex, dense, and cluttered scenes, reflecting its strong generalization capability. \Cref{fig:vis_1} compares RT-DETRv4 with novel real-time detectors, including YOLOv12~\cite{tian2025yolov12}, YOLOv13~\cite{lei2025yolov13}, D-FINE~\cite{peng2025d}, and DEIM~\cite{huang2025deim}.
Across various visual conditions, RT-DETRv4 consistently exhibits superior detection accuracy and detects more objects that are often missed by other methods, underscoring its effectiveness in practical inference.

\Cref{fig:vis_2} compares the deepest features before and after semantic injection, suggesting that the detector learns more robust semantic representations under the guidance of the semantic teacher. 
Before semantic injection, deep features exhibit noticeable noise interference and lack robustness. 
In contrast, our method substantially enhances high-level semantics, leading to improved representations that ultimately benefit detection performance.

\section{Training cost analysis}
\label{sec:training_cost}

We analyze the additional training cost introduced by our method and find that the overhead is modest. All experiments are conducted on NVIDIA GeForce RTX 3090 GPUs, where \textit{Training GPU Hours} denote the accumulated GPU time for the full training schedule and \textit{Memory Per Image} measures the average memory usage per input image. Both the baseline and our method follow the same training schedule for fair comparison. As summarized in \Cref{tab:training_cost}, our method introduces only minor increases in GPU hours and memory across all model scales. For example, in the L model, the total training time increases from 152.47 to 166.68 GPU hours (9.3\%), while the average memory per image increases slightly from 1.523 to 1.535 GiB (0.8\%). This small overhead arises because the semantic teacher is frozen during training and performs only a forward pass, introducing minimal additional computation and memory usage. Importantly, our method introduces zero extra cost during inference and deployment.

%%%%%%%%%%%%%%%%%%%%%%%%%%%%%%%%%%%%%%%%%%%%%%%%
\section{Gradient dynamics under different strategies}
\label{sec:gradients_analysis}
%%%%%%%%%%%%%%%%%%%%%%%%%%%%%%%%%%%%%%%%%%%%%%%%
We visualize the modulation dynamics of GAM under its optimal configuration ($\rho = 2.0\%$, $\delta = 0.1\%$) in~\Cref{fig:gam_dynamics}, showing how it adjusts the AIFI gradients and the DSI loss weights during training. This provides an intuitive illustration of how the GAM strategy operates in practice.
Taking the L model as an example, we compare the average gradients of different configurations, as illustrated in~\Cref{fig:gradient_comparison}. 
Specifically, we analyze the baseline model (DEIM-L~\cite{huang2025deim}), DSI with static weights $\lambda{=}2$ and $\lambda{=}4$ (which perform relatively well in the early training stage), DSI with the best static weight $\lambda{=}20$, and DSI combined with the proposed GAM strategy. 
For each configuration, we report both the average gradients computed every 12 epochs and the overall average gradients across the main modules, as summarized in~\Cref{tab:dynamic_gradients}.

From the comparison, we observe that the baseline AIFI module indeed suffers from insufficient supervision, as its gradient ratio drops sharply after around 10 epochs. 
Given that AIFI produces the deepest and most semantic feature representation in the detector, this phenomenon supports our hypothesis of a semantic bottleneck.
Static-weight DSI partially alleviates this issue but fails to balance the varying supervision strength required across early, middle, and late training phases. 
Even with the introduction of DSI, using static weights cannot achieve stable and efficient semantic injection. 
As shown in the ablation study on the loss weighting strategy of the main paper, smaller $\lambda$ values provide only marginal AP improvements, while excessively increasing $\lambda$ leads to drastic gradient fluctuations in AIFI during the early training stage, sometimes even degrading initial performance and consequently hindering the final results. 

In contrast, the proposed GAM adaptively regulates the distillation strength throughout training. 
The parameter $\rho$ determines the target gradient ratio controlling the injection strength, while $\delta$ controls the tolerance around the target ratio and thus the sensitivity of the modulation. 
Properly tuned values of these parameters enable GAM to surpass all static-weight configurations, achieving both higher detection accuracy and improved training stability.

%%%%%%%%%%%%%%%%%%%%%%%%%%%%%%%%%%%%%%%%%%%%%%%%
\section{Ablation on GAM hyper-parameters}
\label{sec:gam_params}
%%%%%%%%%%%%%%%%%%%%%%%%%%%%%%%%%%%%%%%%%%%%%%%%
Based on the analysis of multiple static weighting settings of the L model~(\textit{cf.}~\cref{sec:gradients_analysis}), we initially determine that the optimal range for $\rho$ lies between 1.0\% and 2.5\%, and for $\delta$ between 0 and 0.5\%. We then perform a comprehensive ablation study on their combinations in~\Cref{tab:ablation_rho_delta} and finally set $\rho = 2.0\%$ and $\delta = 0.1\%$.  
Here, $\rho$ governs the overall magnitude of semantic injection, while $\delta$ controls how smoothly the modulation is introduced during the warm-up phase and how sensitively it adapts during the progressive injection stage.  
These optimal values ensure that the semantic injection is activated in a gradual yet effective manner, achieving a balanced and beneficial modulation strength.

\begin{table}[!t]
\centering
\setlength{\tabcolsep}{4.3pt}
\renewcommand{\arraystretch}{1}
\caption{\textbf{Ablation on GAM hyper-parameters.} DEIM-L is adopted as the baseline, with all models trained for 50 epochs and 8 EMA epochs.}
\label{tab:ablation_rho_delta}
\resizebox{\textwidth}{!}{%
    \begin{tabular}{c|cccccc|cccccc}
        \toprule
        \textbf{$\rho\%,\delta\%$} & 1, 0 & 1.5, 0 & 1.8, 0 & 2, 0 & 2.2, 0 & 2.5, 0 & 1, 0.1 & 1.5, 0.1 & 1.8, 0.1 & \textbf{2, 0.1} & 2.2, 0.1 & 2.5, 0.1\\
        \arrayrulecolor{gray}\midrule
        \textbf{AP$^{val}$} & 55.0  & 55.1 & 55.2 & 55.2 & 55.1 & 55.0 & 55.0  & 55.1 & 55.2 & \textbf{55.4} & 55.2 & 55.1 \\
        \arrayrulecolor{black}\midrule[0.8pt]
        \textbf{$\rho\%,\delta\%$} & 1, 0.25 & 1.5, 0.25 & 1.8, 0.25 & \underline{2, 0.25} & 2.2, 0.25 & 2.5, 0.25 & 1, 0.5 & 1.5, 0.5 & 1.8, 0.5 & 2, 0.5 & 2.2, 0.5 & 2.5, 0.5 \\
        \arrayrulecolor{gray}\midrule
        \textbf{AP$^{val}$} & 54.9  & 55.1 & 55.1 & \underline{55.3} & 55.2 & 55.2 & 54.9 & 55.0 & 55.0 & 55.2 & 55.1 & 55.0 \\
        \arrayrulecolor{black}\bottomrule
    \end{tabular}%
}
\end{table}

\begin{table}[!t]
\centering
\scriptsize
\caption{\textbf{CKA similarity between detector layers and VFM (DINOv3).}}
\label{tab:cka_similarity}
\begin{tabularx}{\textwidth}{>{\raggedright\arraybackslash}p{3.8cm} | >{\centering\arraybackslash}X >{\centering\arraybackslash}X >{\centering\arraybackslash}X >{\centering\arraybackslash}X}
    \toprule
    \textbf{Feature map} & \textbf{$S_3$} & \textbf{$S_4$} & \textbf{$S_5$} & \textbf{$F_5$} \\
    \midrule
    \textbf{CKA similarity} & 0.05 & 0.30 & 0.40 & \textbf{0.59} \\
    \bottomrule
\end{tabularx}
\end{table}

%%%%%%%%%%%%%%%%%%%%%%%%%%%%%%%%%%%%%%%%%%%%%%%%
\section{CKA analysis of semantic injection position}
%%%%%%%%%%%%%%%%%%%%%%%%%%%%%%%%%%%%%%%%%%%%%%%%

To empirically provide geometric analysis for deep layer semantic injection, we employ Centered Kernel Alignment (CKA) to quantify representational similarity between the VFM and detector feature maps. CKA is invariant to orthogonal transformations and isotropic scaling, making it more sensitive to meaningful principal component structures and thus suitable for evaluating semantic alignment across architectures. As shown in~\cref{tab:cka_similarity}, the shallow feature $S_3$ is nearly orthogonal to the VFM space (0.05), indicating a large semantic gap as it still primarily captures low-level details. Enforcing alignment at this stage would disrupt the natural information flow. In contrast, the deepest feature $F_5$, after AIFI self-attention, reaches a similarity of 0.59, suggesting that it already lies close to the high-level semantic manifold and exhibits information structures similar to VFM features.

\section{Clarification of the Information Bottleneck principle}
\label{sec:ib_clarification}
\noindent \textbf{Mutual Information.}
Mutual Information (MI) is a fundamental concept in information theory that measures the dependence between two random variables.
For two discrete random variables $Z$ and $W$ with a joint probability mass function $p(z, w)$ and marginal probability mass functions $p(z)$ and $p(w)$, the mutual information $I(Z; W)$ is defined as:
\begin{equation}
I(Z; W) = \sum_{w \in \mathcal{W}} \sum_{z \in \mathcal{Z}} p(z, w) \log \left( \frac{p(z, w)}{p(z)p(w)} \right)
\label{eq:mutual_information}
\end{equation}
Mutual information can also be expressed in terms of entropy ($H$):
\begin{equation}
I(Z; W) = H(Z) - H(Z|W) = H(W) - H(W|Z)
\label{eq:mutual_information_entropy}
\end{equation}
\noindent \textbf{Information Bottleneck.}
The Information Bottleneck (IB) principle~\cite{tishby2000information} postulates that an optimal representation $R$ must maximally compress the input $X$ to discard irrelevant noise while preserving the maximum possible relevant information about the target $Y$. 
The standard IB objective is formulated as the minimization of the Lagrangian functional:
\begin{equation}
\mathcal{L}_{IB} = I(X; R) - \beta I(R; Y)
\label{eq:information_bottleneck_clarification}
\end{equation}
where $\beta$ controls the trade-off between the compression of the input space and the predictive accuracy for the target domain. In an ideal optimization trajectory, the network undergoes a brief fitting phase where $I(R; Y)$ rapidly increases, followed by a prolonged compression phase where $I(X; R)$ decreases, yielding a representation that generalizes better by discarding spurious correlations.

\end{document}